%% file: main.tex
\documentclass[10pt,twocolumn,letterpaper]{article}

\usepackage[pagenumbers]{cvpr}      %

\input{preamble}

\definecolor{cvprblue}{rgb}{0.21,0.49,0.74}
\usepackage[breaklinks,colorlinks,allcolors=cvprblue]{hyperref}

\def\confName{CVPR}
\def\confYear{2027}

\title{How Local Is Local Enough? \\ Measuring the Linear Regime of Image Encoders}

\author{
Jakub Szymkowiak$^{1,2}$ \qquad
Patryk Bartkowiak$^{1}$ \qquad
Wojciech Pa{\l}ubicki$^{1}$ \qquad
Kamil Adamczewski$^{2,3}$\\[6pt]
{\normalsize $^{1}$Adam Mickiewicz University in Pozna\'{n} \qquad
$^{2}$Akces NCBR \qquad
$^{3}$Wroc{\l}aw University of Science and Technology}
}
\begin{document}
\maketitle

\input{sections/00_abstract}
\input{sections/01_introduction}
\input{sections/02_related_work}
\input{sections/03_method}
\input{sections/04_experiments}

\input{sections/05_conclusion}

\section*{Acknowledgments}
This work was supported in part by the European Union through the EOSC-ARENA project (AI Research Enhancement through Networked Agents), Grant Agreement No.~101292597, and by the PIAST-AI Factory project (PIAST AIF), co-financed by the EuroHPC Joint Undertaking through the European Union's Digital Europe Programme and by Poland as the Participating Country under Grant Agreement No.~101250730.
The authors thank Pozna\'{n}skie Centrum Superkomputerowo-Sieciowe (PCSS) for providing the computational resources used in the experiments.

\section*{LLM Usage Statement}
We used LLM assistants (Claude, Anthropic; ChatGPT, OpenAI) in preparing this paper.
In the writing, these tools were used to shorten and polish the text of all sections, and to check the consistency of the text, the figure captions and the references.
In the research, they were used to assist in writing the experimental code and the code that produced the figures, as well as to find related work to read and cite.
All results and claims were produced and verified by the authors.

{
    \small
    \bibliographystyle{ieeenat_fullname}
    \bibliography{main}
}

\input{sections/appendix/appendix_main}

\end{document}

%% file: preamble.tex
\usepackage{array}
\usepackage{booktabs}
\usepackage[table]{xcolor}
\usepackage{multirow}

\usepackage{mathtools}

\newcommand{\tr}{\intercal}

\newcommand{\R}{\mathbb{R}}
\newcommand{\E}{\mathbb{E}}

\newcommand{\Sph}{\mathbb{S}}

\newcommand{\Jac}{\boldsymbol{\mathrm{J}}}

\newcommand{\xx}{\boldsymbol{x}}

\newcommand{\uu}{{\boldsymbol{u}}}
\newcommand{\ww}{\boldsymbol{w}}
\newcommand{\vv}{\boldsymbol{v}}

\newcommand{\bd}{\boldsymbol{\delta}}

\newcommand{\IS}{\mathcal{I}}
\newcommand{\di}{{d_{\IS}}}

\newcommand{\IZ}{\mathcal{Z}}
\newcommand{\dz}{{d_{\IZ}}}

\DeclarePairedDelimiter{\norm}{\|}{\|}
\newcommand{\emptyNorm}{\norm{\, \cdot \,}}

\DeclarePairedDelimiter{\innerprod}{\langle}{\rangle}
\newcommand{\inner}[2]{\innerprod{ #1 , #2 }}

\definecolor{markRounding}{HTML}{767676}
\definecolor{markODIN}{HTML}{C0289F}
\definecolor{markFGSM}{HTML}{262626}
\definecolor{markSG}{HTML}{C8102E}
\definecolor{markIG}{HTML}{6B7A00}
\definecolor{markFreeAT}{HTML}{7A3FC0}
\definecolor{markMaha}{HTML}{EFEFEF}

\newcommand{\markbullet}[1]{\textcolor{#1}{$\bullet$}}

\DeclareRobustCommand{\bODIN}{\markbullet{markODIN}~ODIN~\cite{liang2018enhancing}}
\DeclareRobustCommand{\bFGSM}{\markbullet{markFGSM}~FGSM at one gray level~\cite{goodfellow2015explaining}}
\DeclareRobustCommand{\bSG}{\markbullet{markSG}~Simon-Gabriel et al.~\cite{simongabriel2019first}}
\DeclareRobustCommand{\bIG}{\markbullet{markIG}~Integrated Gradients (median path segment)~\cite{sundararajan2017axiomatic}}
\DeclareRobustCommand{\bFreeAT}{\markbullet{markFreeAT}~free adversarial training~\cite{shafahi2019adversarial}}

%% file: sections/00_abstract.tex
\begin{abstract}

The input Jacobian of an image encoder provides a local linear approximation of its response to input perturbations, which is routinely used to construct adversarial examples, attribute predictions to pixels, and detect out-of-distribution inputs.
But how accurately does this approximation describe changes of finite magnitude?
In this work, we measure the extent of an image encoder's linear regime – the scale at which its response departs from the linear prediction – and characterize the response beyond that scale.
We find that the linear regime is surprisingly short: the departure occurs at less than one gray level, at scales typically smaller than the 8-bit rounding error, and below what many methods use.
Along the Jacobian's leading singular directions, where the encoder is most sensitive, the departure often occurs below the smallest scale we measure.
Beyond the departure scale, the response follows distinct direction-dependent profiles: along random directions, it first exceeds and then falls below the linear prediction, whereas along the leading singular directions, it falls below from the outset.
These results persist across convolutional and transformer architectures, and across supervised, self-supervised, and vision–language training.

\end{abstract}

%% file: sections/01_introduction.tex
\begin{figure*}[t]
    \centering
    \includegraphics[width=\linewidth]{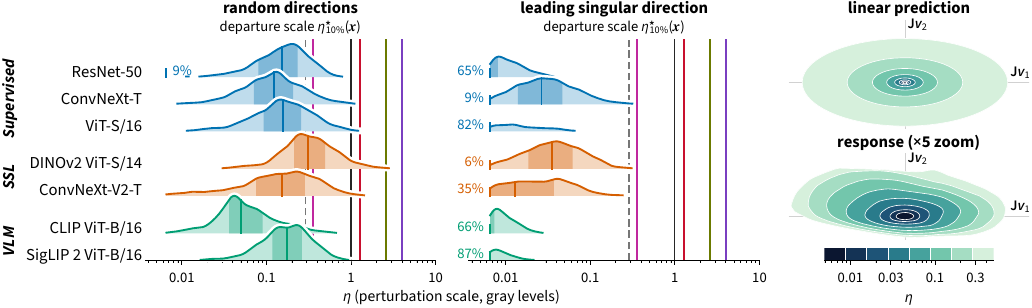}
    \caption{
        \textit{How local is local enough?}
        \textbf{Left:} The linear regime in trained encoders is short and for most images ends below the scales used in: \bODIN, \bFGSM, \bSG, \bIG, \bFreeAT, as well as the 8-bit rounding error (dashed), as shown by the distributions of departure scales at 10\% tolerance (Sec.~\ref{sec:departure-scale}).
        Along the leading singular direction, trained encoders leave their linear regime earlier than along random directions.
        Percentages give the share of images that depart below the smallest scale we measure.
        \textbf{Right:} 
        We measure how quickly an encoder stops following its linearization.
        In the picture, perturbations $\eta \uu$ in the plane spanned by the two leading singular vectors $\vv_1, \vv_2$ of the Jacobian $\Jac$ are predicted to trace ellipses $\eta \Jac \uu$ (Sec.~\ref{sec:setup}), but the measured responses $\Phi(\xx+\eta\uu)-\Phi(\xx)$ grow much more slowly and lose their symmetry about the origin (ResNet-50, a single image, responses projected onto the plane spanned by $\Jac \vv_1$ and $\Jac \vv_2$).
    }
    \label{fig:departure_distributions}
\end{figure*}

\section{Introduction}

Derivatives of an image encoder’s output with respect to its input are used extensively throughout modern computer vision, in methods that construct adversarial examples~\cite{goodfellow2015explaining, moosavidezfooli2016deepfool, khrulkov2018art}, attribute predictions to pixels~\cite{simonyan2014deep}, detect out-of-distribution inputs~\cite{liang2018enhancing}, and quantify sensitivity to input perturbations~\cite{novak2018sensitivity, hoffman2019robust, simongabriel2019first}.\footnote{Most of these methods differentiate a classifier or a loss built on top of an encoder. By the chain rule, such a derivative is the encoder's Jacobian composed with that of the head.}
In each case, they are used to \textit{linearize} the encoder: to replace the nonlinear map it implements with a simpler, linear map $\Jac (\xx)$, called the \textit{input Jacobian}, that predicts the encoder's behavior in the vicinity of an image $\xx$.
For a perturbation $\bd$ of the image, the output of the encoder $\Phi$ changes by approximately $\Jac(\xx) \bd$, with an error that vanishes faster than the perturbation's size $\norm{\bd}$.
Put simply, the linear prediction is accurate \textit{locally}, for perturbations that are \textit{small enough}, and this is how the use of linearization is justified~\cite{simonyan2014deep, hoffman2019robust, khrulkov2018art}.
But how small is small enough?

In practice, often the perturbation size is treated as a hyperparameter, and set to the value that performs best~\cite{liang2018enhancing, lee2018simple, wong2020fast, zhang2024openood}; it is then reported and assumed to be small enough, in most cases without a test.
When it is tested, the test is usually indirect, carried out on the method's own quantity and frequently at a single size, against the method's own standard of agreement~\cite{simongabriel2019first, etmann2019connection, franceschi2018robustness, malo2015geometrical}.
Yet, when performance degrades, the authors of several such methods attribute it, at least in part, to the error of the linear prediction~\cite{dong2018boosting, liu2017delving, tramer2018ensemble, andriushchenko2020understanding, deng2021unified, liang2018enhancing}.

In this work, we propose a framework for measuring how far from an image an encoder's linear prediction remains valid.
As the input moves away from an image along a fixed direction, the encoder follows its linear prediction closely over a span of perturbation sizes, or \textit{scales}, which we call its \textit{linear regime}; the first scale at which it departs, in size or in direction, by more than a set tolerance is its \textit{departure scale}.

We measure the departure scale of convolutional and transformer encoders trained with supervised, self-supervised and vision-language objectives, and find that their linear regimes are short.
The median departure scale along uniformly random directions lies below one gray level, a root-mean-square pixel change of one 8-bit quantization step, in every trained encoder we test, and typically below the scale of the 8-bit rounding error. 
Along the leading singular direction of the Jacobian, the encoder leaves its linear regime at a small fraction of that scale.
For most images, the linear regime ends before the perturbation reaches the scales used in practice (Figure~\ref{fig:departure_distributions}).

To see what these scales mean for an actual method, we measure two published ones, DeepFool~\cite{moosavidezfooli2016deepfool} and ODIN~\cite{liang2018enhancing}, each along its own step direction and at its own step size, and ask whether its steps lie inside the linear regime.
ODIN’s step is larger than the departure scale along its direction, so it acts outside the linear regime, and DeepFool tends to take more iterations where the regime is shorter.

Our contributions are as follows:
\begin{itemize}
    \item We propose a framework for measuring the extent of an encoder's linear regime in terms of its departure scale, the scale at which the encoder's linear prediction ceases to be faithful.

    \item We show that the departure scale is small: below one gray level in trained encoders, and typically below the 8-bit rounding error of an image.

    \item We study the departure scale across directions and training, and characterize the response beyond it.

    \item We apply the framework to published methods, along their own steps, and show that such steps can lie outside the linear regime.
\end{itemize}

%% file: sections/02_related_work.tex
\section{Related Work}

Prior studies of how far a neural network remains faithful to its local linearization have focused on its loss function and activation patterns.
\citet{qin2019adversarial} penalize the residual of the first-order expansion of the loss, and \citet{moosavidezfooli2019robustness} penalize the curvature of the loss along the sign of the gradient, both to improve adversarial robustness.
The linearization of a piecewise-affine network is exact within an activation region; \citet{hanin2019complexity, hanin2019deep} bound the number and density of these regions, while \citet{humayun2024deep} track how their boundaries near the training data move as learning progresses.
\citet{zhang2020empirical} find that the gradient changes little between neighboring regions, so the linear regime can extend across many of them.

A few measurements come closer to ours.
\citet{rifai2011contractive} plot contraction curves: the average change in the features per unit change in the input, measured over spheres of increasing radius.
In contractive autoencoders, they observe a rise to a peak, followed by a decay that they attribute to saturation of the sigmoid units.
We find the same shape in trained image encoders without a contractive penalty or sigmoid units.
\citet{roth2020adversarial} measure the Taylor residual of the penultimate features as a function of distance along a random and an adversarial direction.
The residual does not tell whether the features moved too far or in the wrong direction.
We distinguish these departures through the size ratio and mean cosine, and average the response over a distribution of directions.
\citet{kim2024curved} record the angle between successive displacements of the penultimate feature along a straight line, which compares steps with one another rather than with a linear prediction.
None of these three studies specifies a tolerance, so none gives a scale at which the linear prediction ceases to be valid.
\citet{philipp2021nonlinearity} gives the closest definition, the share of a random offset over which the output change stays within a factor of two of the Jacobian's prediction, measured for fully connected networks.
Our departure scale is defined per image and for any distribution of directions, and we measure it on image encoders, in gray levels.

Among applications that take the linearization as given, those closest to our work use the Jacobian's norm~\cite{novak2018sensitivity, sokolic2017robust, hoffman2019robust, rifai2011contractive} and singular directions.
\citet{berardino2017eigen} use the Jacobian's singular directions as eigendistortions, the perturbations along which the response is predicted to change most and least.
Across a set of models, \citet{feather2025discriminating} define principal distortions, which characterize where the models disagree.
Where the Jacobian at an input stands in for the encoder around it, we measure how far that approximation holds along random directions and along these singular directions.

%% file: sections/03_method.tex
\section{Method}
\label{sec:method}

\subsection{Setup}
\label{sec:setup}
Let $\Phi \colon \R^\di \to \IZ = \R^\dz$ be an image encoder mapping images $\xx \in \IS = [0,1]^\di$ to embeddings $\Phi(\xx) \in \IZ$, with $\di = 3 H W$ for RGB images of resolution $H \times W$, and let $\emptyNorm$ be the usual Euclidean norm on $\IZ$.
We assume that $\Phi$ is differentiable at $\xx$, which holds at almost every input for the networks we consider.
Each model's own input normalization is treated as a part of $\Phi$.

A \textit{perturbation} maps an image $\xx$ to $\xx + \eta \uu$, without clipping the result to $\IS$.
We call its magnitude $\eta > 0$ the \textit{scale}, and the unit vector $\uu \in \Sph^{\di - 1}$ its \textit{direction}.
The \textit{input Jacobian} of $\Phi$ at $\xx$ is the matrix $\Jac (\xx) \in \R^{\dz \times \di}$ with entries $\partial_j \Phi_i (\xx)$.
When the image is fixed, we write $\Jac = \Jac (\xx)$.

\subsection{Size ratio and cosine}
Fix an image $\xx \in \IS$ and a perturbation of scale $\eta$ and direction $\uu$.
We define the encoder's \textit{response} to the perturbed image as the displacement of its embedding, $\Delta_\eta (\xx, \uu) = \Phi(\xx + \eta \uu) - \Phi(\xx)$.
For a small enough scale $\eta$, this response is accurately predicted by the input Jacobian: differentiability of the encoder at $\xx$ guarantees that $\Delta_\eta(\xx, \uu) = \eta \Jac \uu + o(\eta)$, meaning that $\Delta_{\eta} / \eta \to \Jac \uu$ as $\eta \to 0$.
Our measurement compares $\Delta_{\eta} (\xx, \uu) / \eta$ with its \emph{linear prediction} $\Jac \uu$ over increasing finite scales $\eta > 0$ (see Figure~\ref{fig:departure_distributions}, right panel).
Since the comparison depends on the choice of direction, we let $\uu$ follow a law $\mu$ on the unit sphere $\Sph^{\di - 1}$.
Because both $\Delta_{\eta} (\xx, \uu)$ and $\Jac \uu$ are vectors, we compare their lengths and directions separately, using expectations over this law.

For a given law $\mu$, we define the \textit{size ratio} $r_\eta(\xx)$ as the mean squared length of the response per unit scale divided by the mean squared length of its linear prediction, $m_0(\xx) = \E_{\uu \sim \mu} \norm{ \Jac \uu }^2$:
\[
    r_\eta (\xx) = \frac
    {
        \E_{\uu \sim \mu} \norm{ \Delta_{\eta}(\xx, \uu) }^2 / \eta^2
    }
    {
        \E_{\uu \sim \mu} \norm{ \Jac \uu }^2
    } \in [0, +\infty) \, ,
\]
and the \textit{mean cosine} $c_\eta(\xx)$ as the cosine between the response and its linear prediction, averaged over $\mu$:
\[
    c_\eta (\xx) = \E_{\uu \sim \mu}
    \frac
    {
        \inner{ \Delta_{\eta}(\xx, \uu) }{ \Jac \uu }
    }
    {
        \norm{ \Delta_{\eta}(\xx, \uu) } \, \norm{ \Jac \uu }
    } \in [-1, 1] \, .
\]
Both are defined when $\Jac\uu$ and $\Delta_\eta(\xx, \uu)$ are nonzero for $\mu$-almost every $\uu$.
Since both $r_\eta (\xx)$ and $c_\eta (\xx)$ approach $1$ as $\eta \to 0$, we set $r_0 = c_0 = 1$, and refer to these limits as the linear predictions of $r_\eta (\xx)$ and $c_\eta (\xx)$, respectively.
The response is \textit{superlinear} at scale $\eta$ if the embedding moves farther than the Jacobian predicts, i.e., $r_\eta(\xx) > 1$, and \textit{sublinear} if it moves less far, i.e., $r_\eta(\xx) < 1$.
A value of $c_\eta (\xx) < 1$ means that the response turns away from the direction of its linear prediction.

\subsection{Departure scale}
\label{sec:departure-scale}
To summarize the profiles of $r_\eta (\xx)$ and $c_\eta(\xx)$ by a single number per image, we record the first scale at which either of them moves away from its linear prediction by more than a given tolerance.
For a chosen tolerance $\epsilon \in (0,1)$, we define the \textit{departure scale} $\eta^*_{\epsilon}$ as
\[
    \eta^*_\epsilon(\xx) = \inf \left\{ \eta > 0 \;\middle|\;
    \begin{aligned}
        & r_\eta(\xx) \notin \left[ 1 - \epsilon, 1/(1 - \epsilon) \right] \\
        & \text{or } c_\eta(\xx) < 1 - \epsilon
    \end{aligned}
    \right\} .
\]
The interval $[1 - \epsilon, 1/(1-\epsilon)]$ used for $r_\eta$ is symmetric on a logarithmic scale, so a superlinear and a sublinear departure by the same factor count equally.
Both $r_\eta$ and $c_\eta$ are averages over $\mu$, so some directions may depart earlier: the departure scale is an upper bound on the scale up to which every direction stays within the tolerance.

\begin{figure*}[t]
    \centering
    \includegraphics[width=\linewidth]{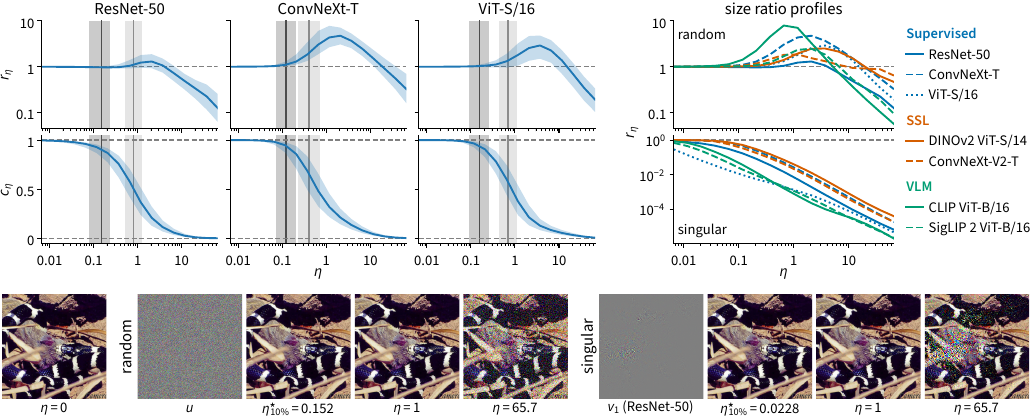}
    \caption{
        Size ratio and cosine profiles of trained encoders, with example perturbations along a random and a singular direction.
        \textbf{Left:} median size ratio $r_\eta$ (top) and mean cosine $c_\eta$ (bottom) of three encoders along uniformly random directions, over 1000 ImageNet validation images.
        Vertical lines are medians of the departure scales $\eta^*_{10\%}$ (dark) and $\eta^*_{50\%}$ (light), with shaded regions their interquartile ranges.
        \textbf{Right:} median size ratio $r_\eta$ of seven encoders – profiles along uniform directions (top) first turn superlinear, while those along the leading singular direction (bottom) are sublinear from the start.
        \textbf{Bottom:} an example image ($\eta = 0$) and its perturbations along a single random direction $\uu$ and along a leading singular direction $\vv_1$ of ResNet-50, at the image's own departure scale $\eta^*_{10\%}$ and at larger scales; the $\uu$ and $\vv_1$ tiles show the directions themselves, rescaled to the full gray range.
    }
    \label{fig:measurement-profiles}
\end{figure*}

\subsection{Direction laws}
\label{sec:laws}
The law $\mu$ describes the directions along which the linear prediction is tested, and so far we left it unspecified.
In practice, the directions come from the method that uses the Jacobian, and we then take $\mu$ as the method's own distribution; for a single direction $\uu$, this is the point mass $\mu = \delta_{\uu}$.
Below, we outline two simple laws that allow for a method-agnostic measurement, each of which can be viewed as the uniform law on the unit sphere of a linear subspace of the perturbation space $\R^{\di}$.
A third such law, supported on the kernel of $\Jac$, is described in Appendix~\ref{app:kernel}.

\paragraph{Uniform.}
Under the uniform law $\mu := \mathrm{Unif}(\Sph^{\di - 1})$, all directions are weighted equally.
In this case, $\E_{\uu \sim \mu} [\uu \uu^\tr] = \boldsymbol{I} / \di$, so $m_0(\xx) = \norm{ \Jac }_F^2 / \di$, with $\norm{\cdot}_F$ the Frobenius norm.
The norm $\norm{ \Jac }_F$ serves as a first-order proxy for the change of the embedding under small random perturbations, both as a sensitivity measure~\cite{novak2018sensitivity} and as a regularizer~\cite{hoffman2019robust}.
The size ratio under the uniform law shows at which scales this proxy holds, and by what factor it is off elsewhere.

\paragraph{Singular.}
Let $\vv_1, \dots, \vv_{\dz} \in \R^{\di}$ denote the right singular vectors of $\Jac$ and $\lambda_1 \ge \dots \ge \lambda_{\dz} \ge 0$ the corresponding squared singular values, $\lambda_k = \norm{ \Jac \vv_k }^2$.
Since a singular vector is defined only up to sign, for each rank $k$ with $\lambda_k > 0$ we define the law $\mu_{(k)} := (\delta_{\vv_k} + \delta_{-\vv_k}) / 2$.
We then have $m_0(\xx) = \lambda_k$.
For $k = 1$, this is the squared spectral norm $\lambda_1 = \norm{ \Jac }_2^2$, which serves as a first-order proxy for the largest change a perturbation of a given magnitude can cause~\cite{sokolic2017robust, simongabriel2019first}.
The size ratio under $\mu_{(1)}$ then shows at which scales this proxy holds along $\vv_1$, and increasing $k$ extends the comparison to directions of progressively lower sensitivity.

\subsection{Velocity along the ray}
\label{sec:velocity}
Finally, the response $\Delta_\eta(\xx, \uu)$ is determined by the Jacobian along the whole ray from $\xx$ to $\xx + \eta\uu$, since for locally Lipschitz encoders the fundamental theorem of calculus gives
\[
    \Delta_\eta(\xx, \uu) = \int_0^\eta \Jac(\xx + t\uu)\uu \, \mathrm{d}t .
\]
Call the integrand $\ww_\uu(t) = \Jac(\xx + t\uu)\uu$ the \textit{velocity} of the embedding along the ray $t \mapsto \xx + t\uu$; at $t = 0$ it is the linear prediction $\Jac\uu$.
Let
\[
    m_\eta(\xx) = \frac{1}{\eta} \int_0^\eta \E_{\uu \sim \mu} \norm{\ww_\uu(t)}^2 \, \mathrm{d}t
\]
be the mean squared speed along the ray, which tends to $m_0(\xx)$ as $\eta \to 0$.
The size ratio then splits as $r_\eta = \sigma_\eta \, \tau_\eta$, into the \textit{speed factor} and \textit{steadiness}
\[
    \sigma_\eta(\xx) = \frac{m_\eta(\xx)}{m_0(\xx)} ,
    \qquad
    \tau_\eta(\xx) = \frac{\E_{\uu \sim \mu} \norm{\Delta_\eta(\xx, \uu)}^2 / \eta^2}{m_\eta(\xx)} .
\]
By the Cauchy--Schwarz inequality $\tau_\eta \le 1$, with equality only if the velocity is constant along the ray.
The response is therefore superlinear at scale $\eta$ only if $\sigma_\eta > 1$, and sublinear if the velocity keeps its starting speed and only turns.

%% file: sections/04_experiments.tex
\section{Experiments}
\label{sec:experiments}

We apply the measurement framework of Section~\ref{sec:method} to image encoders.
The seven main ones are three supervised (ResNet-50, ConvNeXt-T, ViT-S/16), two self-supervised (DINOv2 ViT-S/14, ConvNeXt-V2-T) and two vision--language (CLIP ViT-B/16, SigLIP~2 ViT-B/16); Table~\ref{tab:models} lists all encoders we consider and specifies the embedding $\Phi(\xx)$ for each.
Unless stated otherwise, we measure a fixed, class-balanced set of 1000 ImageNet validation images.

We report scales in gray levels: a perturbation of scale $\eta$ changes a pixel value by $\eta / \sqrt{\di}$ in root mean square, which is $255 \, \eta / \sqrt{\di}$ on the $0$ to $255$ scale.
One gray level is a single step of 8-bit quantization, and corresponds to $\eta \approx 1.52$ at ImageNet resolution ($\di = 3 \cdot 224^2$).
We use 17 perturbation scales, with $\eta$ growing from $0.01$ to $100$ by a factor of $10^{1/4}$ per step, that is, from $0.0066$ to $66$ gray levels at ImageNet resolution.

Under the uniform law, we draw 2048 random directions per image, as 1024 antipodal pairs $\pm \uu$.
We report the departure scale at a tolerance of $\epsilon = 10\%$, written $\eta^*_{10\%}$, and at $50\%$ where stated.
Appendix~\ref{app:measurement-details} reports the bootstrap precision of $\eta^*$, validates the procedure on synthetic encoders whose departure scales are known exactly, and describes the rules by which we exclude scales dominated by floating-point error.
Appendices~\ref{app:other-image-sets} and~\ref{app:other-layers} repeat the uniform-law measurement on three other image sets and at other layers of the supervised encoders.

\begin{figure}
    \centering
    \includegraphics[width=1\linewidth]{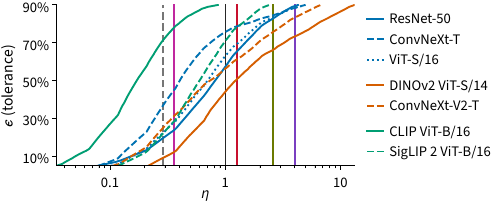}
    \caption{
        Median departure scale $\eta^*_{\epsilon}$ under the uniform law, with reference scales as in Fig.~\ref{fig:departure_distributions} (vertical lines).
        Reaching one gray level takes $\epsilon > 50\%$ in all encoders except DINOv2.
    }
    \label{fig:tolerance}
\end{figure}

\subsection{The extent of the linear regime}
\label{sec:extent}

For all trained encoders (Supervised, SSL, and VLM; the adversarially trained encoders are treated separately), at least $94\%$ of the departure scales $\eta^*_{10\%}$ under the uniform law lie below one gray level (Fig.~\ref{fig:departure_distributions}, random).
The median departure scale is below a third of a gray level in each of them, ranging from $0.05$ gray levels for CLIP to $0.31$ for DINOv2, with the three supervised encoders between $0.12$ and $0.16$ (Table~\ref{tab:radii_uniform}).
Depending on the encoder, the mean cosine (ResNet-50, SigLIP-2) or the size ratio (the other five) leaves the tolerance at the smaller median scale.
At a 50\% tolerance, the median departure scale ranges from $0.17$ gray levels for CLIP to $1.23$ for DINOv2, and lies below one gray level in six of the seven encoders (Table~\ref{tab:radii_uniform}; Fig.~\ref{fig:measurement-profiles}, left, for the supervised encoders; Fig.~\ref{fig:tolerance} for all tolerances).

Along the leading singular direction, the linear prediction moves the embedding $\sim$130--220 times farther than along a random direction, for a perturbation of the same magnitude, and the departure occurs at smaller scales.
In every trained encoder, at least $90\%$ of the images have a departure scale $\eta^*_{10\%} < 0.1$ gray levels.
In ResNet-50, ViT-S/16, CLIP and SigLIP-2, the median departure scale lies at or below the smallest scale we measure, $0.0066$ gray levels, and in the remaining three encoders it is 4 to 12 times shorter than along random directions (Fig.~\ref{fig:departure_distributions}, singular; Table~\ref{tab:radii_rank}).

\paragraph{Comparison with 8-bit rounding.}
A finer reference than one gray level is the scale of the image's \textit{8-bit rounding error}, $1 / \sqrt{12} \approx 0.29$ gray levels under the uniform quantization-noise model  \cite{bennett1948spectra} (Fig.~\ref{fig:departure_distributions}, dashed line).
The median departure scale $\eta^*_{10\%}$ along random directions lies below this value in every trained encoder except DINOv2, whose median is $0.31$ gray levels, and along the leading singular direction in every encoder.

\begin{figure}
    \centering
    \includegraphics[width=1\linewidth]{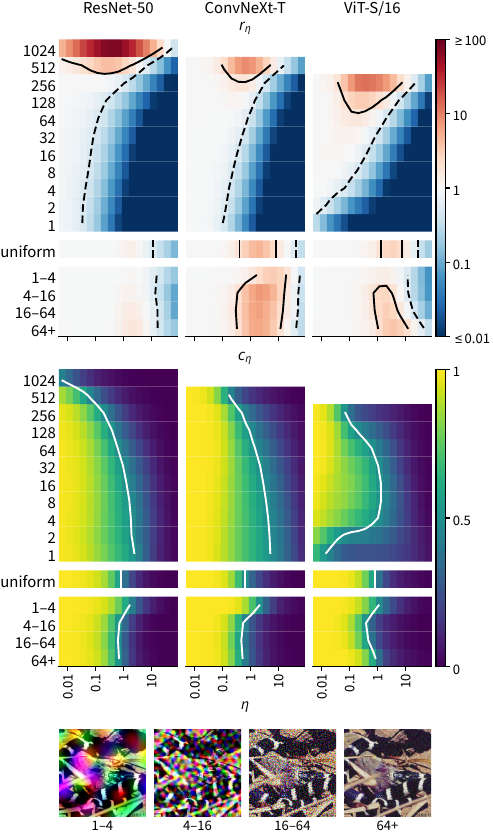}
    \caption{
       Size ratio $r_\eta$ (top) and mean cosine $c_\eta$ (bottom) of three supervised encoders across direction laws, as medians over images.
       Each row is one law: the singular ranks $k$, the uniform law, and the uniform law restricted to four ranges of spatial frequency (cycles per image); blank ranks have no positive singular value.
       Blue is sublinear ($r_\eta < 1$), red superlinear ($r_\eta > 1$); lines mark $r_\eta = 2$ (solid), $r_\eta = 1/2$ (dashed) and $c_\eta = 0.5$ (white).
       Leading ranks turn sublinear and trailing ranks superlinear, at a transition rank that differs between encoders.
       In every frequency range the size ratio rises above $1$, as under the uniform law, and least in the lowest range.
    }
    \label{fig:singular-freq-heatmaps}
\end{figure}

\subsection{Beyond the linear regime}
At larger scales, what the response does instead depends on the direction of the perturbation.

\begin{figure}
    \centering
    \includegraphics[width=\linewidth]{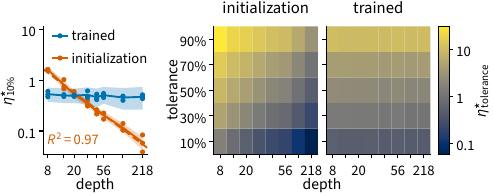}
    \caption{
        Departure scale $\eta^*$ of CIFAR-10 ResNets~\cite{he2016deep} against depth, at initialization and after training, under the uniform law.
        \textbf{Left:} median $\eta^*_{10\%}$ per seed (points) and their mean over seeds (lines), with interquartile ranges shaded; the dashed line is a power-law fit at initialization, with exponent $-0.98$ and 95\% confidence interval $[-1.06, -0.91]$.
        \textbf{Right:} median departure scale at tolerances from 10\% to 90\%.
        At initialization, $\eta^*_{10\%}$ falls in inverse proportion to depth, while after training the departure scale is nearly constant across depth at every tolerance.
    }
    \label{fig:depth-family-rn}
\end{figure}

\paragraph{Profiles.}
Along uniformly random directions, the median size ratio $r_\eta$ departs upward in all seven trained encoders (Fig.~\ref{fig:measurement-profiles}, right).
It peaks between $0.66$ and $3.7$ gray levels, at values from $1.3$ for ResNet-50 to $8.0$ for CLIP, and returns below $1$ between $4$ and $20$ gray levels.
The median profile thus shows a superlinear response at intermediate scales and a sublinear one at larger scales.
The median of the mean cosine $c_\eta$ falls below $0.5$ before one gray level in six of the seven encoders, and at $1.7$ gray levels in DINOv2.
Most single images show the same rise.
For ResNet-50, the size ratio rises above the tolerance in $70\%$ of individual images, and for each of the other encoders in at least $88\%$.

Along the leading singular direction, the median size ratio never exceeds $1$ over the measured scales and departs downward in every encoder.
At $1.17$ gray levels, the measured scale nearest to one gray level, it lies between $0.0006$ for CLIP and $0.032$ for DINOv2, that is, at most $3.2\%$ of the predicted squared length.

\paragraph{Singular ranks.}
To see how the response changes across the singular directions of the Jacobian, we measure the profiles of ResNet-50, ConvNeXt-T and ViT-S/16 at ranks $k = 1, 2, 4, \dots$ (Fig.~\ref{fig:singular-freq-heatmaps}).
At the leading ranks the median size ratio departs downward, and at the trailing ranks upward.
The transition occurs between ranks $256$ and $512$ for ResNet-50, $64$ and $128$ for ConvNeXt-T, and $16$ and $32$ for ViT-S/16, and beyond it the peak of the median size ratio grows with the rank (Appendix~\ref{app:beyond-linear-regime}).
The response thus falls short of its linear prediction along the most sensitive singular directions, and exceeds it along the less sensitive ones.

\paragraph{Spatial frequency.}
A uniformly random direction has, in expectation, the same energy in every Fourier component, and about three quarters of the components lie above $64$ cycles per image.
To see whether the short linear regime and the superlinear response persist for smoother perturbations, we restrict the uniform law to one of four ranges of radial spatial frequency, $1$--$4$, $4$--$16$, $16$--$64$ and above $64$ cycles per image, for the same three encoders (Fig.~\ref{fig:singular-freq-heatmaps}).
In every range, the median departure scale $\eta^*_{10\%}$ lies between $0.07$ and $0.40$ gray levels, and the median size ratio exceeds $1$, but its peak is smallest in the lowest range (Appendix~\ref{app:beyond-linear-regime}).
The short linear regime persists even for low-frequency perturbations.

\begin{figure*}[t]
    \centering
    \includegraphics[width=\linewidth]{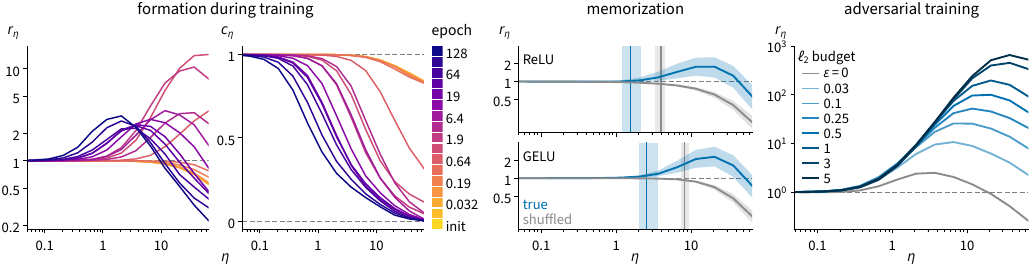}
    \caption{
        Effect of training on the profiles under the uniform law, as medians over images.
        \textbf{Left:} size ratio $r_\eta$ and mean cosine $c_\eta$ of ConvNeXt-T across ImageNet training checkpoints.
        \textbf{Middle:} $r_\eta$ of CIFAR-10 ResNet-18 with ReLU and GELU activations, trained on true and on shuffled labels; shaded regions are interquartile ranges, and vertical lines the median $\eta^*_{10\%}$.
        \textbf{Right:} $r_\eta$ of ResNet-50 trained adversarially at increasing $\ell_2$ budgets~\cite{salman2020adversarially}.
        The superlinear response forms early during training, is absent under shuffled labels, and grows with the adversarial budget.
    }
    \label{fig:training}
\end{figure*}

\begin{figure*}[t]
    \centering
    \includegraphics[width=\linewidth]{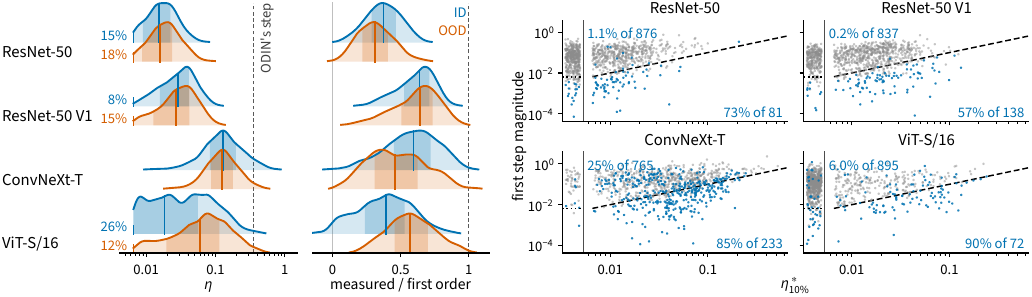}
    \caption{
        ODIN~\cite{liang2018enhancing} and DeepFool~\cite{moosavidezfooli2016deepfool} along their own steps, on four ImageNet classifiers.
        \textbf{Left:} ODIN on ImageNet validation images (ID) and OpenImage-O images (OOD), as distributions with medians and interquartile ranges.            The first column shows the departure scale $\eta^*_{10\%}$ along ODIN's direction, with the share of images at or below the smallest measured scale as a percentage.
        The second column shows the measured change of $\log S$ (Sec.~\ref{sec:odin-deepfool}) at ODIN's step relative to its first-order prediction.
        \textbf{Right:} DeepFool's first-step magnitude against the departure scale $\eta^*_{10\%}$ along its direction.
        Each point is an image, blue if its predicted label changes after the step.
        Above the dashed diagonal, the step is longer than the departure scale; percentages give the share of images on either side whose label changes.            Images at or below the smallest measured scale are in the strip on the left, where the dotted line continues the diagonal.
        Strip images below the dotted line have a step that cannot be compared with their departure scale, and are excluded from both counts.
    }
    \label{fig:deepfool}
\end{figure*}

\subsection{The effect of training}
\label{sec:effect-of-training}
\paragraph{Initialization.}
At random initialization, the median departure scale $\eta^*_{10\%}$ under the uniform law is $0.26$ gray levels for ResNet-50, $13.2$ for ConvNeXt-T and $6.9$ for ViT-S/16, against $0.15$, $0.12$ and $0.16$ after training (Appendix~\ref{app:effect-of-training}).
The three scales thus differ by a factor of about $50$ at initialization, and by $1.3$ after training.
In CIFAR-10 ResNets of depth $8$ to $218$, the departure scale likewise varies with depth at initialization and is nearly constant after training (Fig.~\ref{fig:depth-family-rn}).
In both cases, training brings architectures with very different linear regimes at initialization to a common, short one.
The median size ratio of the untrained ResNet-50 and ViT-S/16 rises without reaching a peak within the measured scales, while that of the untrained ConvNeXt-T only falls.

\paragraph{Formation during training.}
We follow ConvNeXt-T over $128$ epochs of ImageNet training, measuring its response on $256$ validation images (Fig.~\ref{fig:training}, left).
During the first third of an epoch, the median size ratio departs downward and the median departure scale stays above $13$ gray levels.
Between epochs $0.32$ and $0.64$, the share of images whose size ratio departs upward rises from $16\%$ to $93\%$, while the median departure scale falls from $16.2$ to $5.1$ gray levels.
By the end of the first epoch it has fallen to $1.4$ gray levels, and over the remaining training it reaches $0.17$.
A peak of the median size ratio then appears within the measured scales and moves inward, to $2.1$ gray levels by the final epoch.

\paragraph{Memorization.}
For ResNet-18 trained on CIFAR-10 with true labels, the size ratio rises above the tolerance in $85\%$ of $1000$ test images with ReLU activations, and in $92\%$ with GELU (Fig.~\ref{fig:training}, middle).
With shuffled labels, the size ratio rises above the tolerance in none of the images, and the median departure scales are $2.5$ and $3.2$ times as large, respectively.
The upward response is therefore not an automatic consequence of training a nonlinear network on image data; the label assignment matters.

\paragraph{Adversarial training.}
For ResNet-50 trained adversarially with $\ell_2$ budgets from $0.03$ to $5$~\cite{salman2020adversarially}, the median departure scale remains between $0.20$ and $0.23$ gray levels, close to its value without adversarial training (Fig.~\ref{fig:training}, right).
Beyond the departure scale, the peak of the median size ratio grows with the budget by more than two orders of magnitude, from $2.6$ without adversarial training to $658$ at the largest budget, and moves from $3.7$ to $37$ gray levels.
Over the same budgets, the median of $m_0$, the mean squared length of $\Jac\uu$, decreases by a factor of about $1700$, so the growing ratio is measured against a shrinking linear prediction.
Adversarial training shrinks the linear prediction without extending the scale over which it holds.
Appendix~\ref{app:effect-of-training} reports results for $\ell_\infty$ budgets.

\subsection{Published methods along their own steps}
\label{sec:odin-deepfool}
The direction laws considered so far are independent of any particular method.
We now measure two published methods, ODIN~\cite{liang2018enhancing} and DeepFool~\cite{moosavidezfooli2016deepfool}, each with the law $\mu$ placed on the direction of its own step (Sec.~\ref{sec:laws}), and compare the scale of that step with the departure scale along it.
We use four ImageNet classifiers, each consisting of an encoder and a linear head, namely ResNet-50, ResNet-50~V1 (the OpenOOD benchmark model~\cite{zhang2024openood}), ConvNeXt-T and ViT-S/16.
For ODIN, the size ratio and the cosine are averaged over the two signs of that direction, and for DeepFool they are taken along the step's own sign.

\paragraph{ODIN.}
ODIN detects out-of-distribution images using $S$, the largest softmax probability at temperature $T$.
Before scoring, it perturbs the image by one step of fixed scale along the sign of the gradient of $\log S$, a step motivated by the first-order expansion of $\log S$.
We follow OpenOOD's ImageNet configuration, with $T=1000$ and a step of $0.357$ gray levels, on the $1000$ ImageNet validation images and on $1000$ images from OpenImage-O~\cite{yang2022openood}.
At this temperature, the departure of $\log S$ from first order comes from the encoder (Appendix~\ref{app:deepfool-odin}).

Along ODIN's direction, the median departure scale $\eta^*_{10\%}$ ranges from $0.015$ to $0.13$ gray levels across the four models and the two image sets (Fig.~\ref{fig:deepfool}, left).
ODIN's step is longer than the departure scale in every image for the two ResNets and in at least $93\%$ of the images for the other two models, by a median factor of $2.8$ to $24$.
At the step, the median size ratio lies between $0.21$ and $0.68$, and the change of $\log S$ is, in the median, $31\%$ to $64\%$ of its first-order prediction (Fig.~\ref{fig:deepfool}, left, second column).
The change has the predicted sign in at least $97.8\%$ of the images.
The first-order prediction thus generally gives the sign of the score's change at ODIN's step, but overestimates its size.

\paragraph{DeepFool.}
DeepFool searches for a perturbation that changes the predicted label, by repeatedly linearizing the classifier and stepping to the nearest class boundary of the linearization, with an overshoot of $2\%$.
If the response followed its linear prediction, the first step would change the label in every image.
It does so in $8.5\%$ to $39.4\%$ of the images across the four models.
At that step, the score difference has moved by $66\%$ to $94\%$ of its predicted change in the median.

The first step has a median scale of $0.06$ to $0.13$ gray levels, below the 8-bit rounding error (Sec.~\ref{sec:extent}), while the median departure scale $\eta^*_{10\%}$ along it is at most $0.04$ gray levels.
In $77\%$ to $90\%$ of the images, the step is longer than the departure scale (Fig.~\ref{fig:deepfool}).
Among these images, it changes the label in $0.2\%$ to $6.0\%$ for three models, and in $25.5\%$ for ConvNeXt-T.
For steps not longer than the departure scale, the shares range from $57\%$ to $90\%$.
A step longer than the departure scale thus mostly leaves the label unchanged, while a step within it does not guarantee a change.

The number of DeepFool iterations tends to increase with the ratio of the first-step scale to the departure scale.
Across the four models, the median iteration count is one when the first step is not longer than the departure scale, two when it is up to three times as long, and three or four when it is more than ten times as long.
In the median, the final perturbation is within $0.6\%$ of the shortest perturbation along its direction that changes the label.

\begin{figure}
    \centering
    \includegraphics[width=\linewidth]{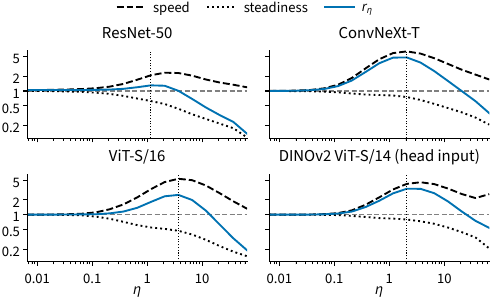}
    \caption{
        The size ratio $r_\eta$ (solid) of four encoders along uniformly random directions as the product of the speed factor $\sigma_\eta$ (dashed) and steadiness $\tau_\eta$ (dotted), as geometric means over 250 ImageNet validation images.
        Vertical lines correspond to the peak of $r_\eta$.
        The size ratio rises with the speed factor, and returns below $1$ as steadiness falls, while the speed factor is still above $1$.
    }
    \label{fig:velocity}
\end{figure}

\subsection{Speed and steadiness along the ray}
\label{sec:velocity-exp}
We measure the velocity under the uniform law at $t = 0$ and at the 17 perturbation scales, for 250 of the images and 256 directions with their antipodes, in ResNet-50, ConvNeXt-T, ViT-S/16 and DINOv2 (head input, Table~\ref{tab:models}).
Figure~\ref{fig:velocity} shows the resulting split $r_\eta = \sigma_\eta \, \tau_\eta$ of Section~\ref{sec:velocity}, as geometric means over images.

In all four encoders, the speed factor $\sigma_\eta$ rises to a peak of $2.3$ to $6.3$, at $2.1$ or $3.7$ gray levels, and remains above $1$ up to the largest measured scale.
Over the same scales, steadiness $\tau_\eta$ decreases to between $0.11$ and $0.21$.
The velocity also turns away from its starting direction, as measured by its inner product with the starting velocity, averaged over directions and normalized by the two root-mean-square speeds.
This inner product is below $0.5$ from about $0.3$ gray levels on in the three supervised encoders and from $0.6$ on in DINOv2, and below $0.1$ beyond $1.0$ to $2.7$ gray levels.

The size ratio initially rises as $\sigma_\eta$ increases, and falls below $1$ as $\tau_\eta$ decreases.
Its median returns below $1$ between $4$ and $26$ gray levels, where the normalized inner product is below $0.1$ in all four encoders.
Among the images whose size ratio rises above $1$ and subsequently returns below it, $199$ to $233$ of the $250$, $\sigma_\eta$ is still above $1$ at that return in at least $99\%$, with a median of $2.2$ to $2.7$.
The response there falls below its linear prediction while $m_\eta$ still exceeds $m_0$; the velocity has by then become less steady and less aligned with its starting value.

%% file: sections/05_conclusion.tex
\section{Conclusion}

Our results put a scale on how small an image perturbation has to be for a trained encoder's linear prediction to remain faithful, namely a fraction of a gray level, and typically below the 8-bit rounding error.
Several widely used methods operate above that scale, where this cannot be assumed without a test.
While a method may work where its linear prediction is no longer faithful, the prediction's validity cannot be inferred from that success.

What determines the extent of the linear regime and the response beyond it is not yet fully understood. 
Although our framework applies to other models with continuous inputs and outputs, whether their linear regimes are similarly short remains to be measured.

%% file: sections/appendix/appendix_main.tex
\clearpage
\setcounter{footnote}{0}
\maketitlesupplementary
\appendix
\input{sections/appendix/A}
\input{sections/appendix/B}
\input{sections/appendix/C}

%% file: sections/appendix/A.tex
\section{Measurement details}
\label{app:measurement-details}

We give the sampling and numerical details of the measurements in Section~\ref{sec:experiments} and describe the encoders used.

\paragraph{Images.}
The 1000 ImageNet validation images contain one image per class, the one whose file name comes first among the validation images of that class.
Each image is resized to $256$ pixels along its shorter side, center-cropped to $224 \times 224$, and scaled to $[0, 1]$.
We add the perturbation before applying the model-specific input normalization, which is part of $\Phi$.
For ODIN (Section~\ref{sec:odin-deepfool}), we also use $1000$ images sampled at random from the $15{,}869$ OpenImage-O test images of OpenOOD~v1.5~\cite{zhang2024openood}, with the same preprocessing.
The CIFAR-10 images are used at their native resolution of $32 \times 32$, where $\di = 3072$ and one gray level corresponds to $\eta \approx 0.22$.
We adjust the $17$ scales to cover the same range in gray levels as for ImageNet.

\paragraph{Clipping.}
We do not clip perturbed images to $[0, 1]$.
For uniformly random perturbations, clipping would affect few pixel values at the smaller departure scales we report.
Pixel values are multiples of $1/255$, and for a uniformly random perturbation the largest absolute pixel change is typically about $5$ times the root-mean-square change in gray levels (Appendix~\ref{app:reference-scales}).
Below roughly $0.2$ gray levels, we expect only values already at $0$ or $1$ to leave the range.
These boundary values account for $0.03\%$ of pixel values in the median image and less than $0.5\%$ in three quarters of the images.
At larger scales, including those near the peak of the size ratio, more pixel values would be affected by clipping. We have not measured its effect at these scales.

\subsection{Sampling details}
\label{app:sampling}

\paragraph{Taylor expansion.}
Suppose $\Phi$ is smooth in a neighborhood of $\xx$.
The second-order Taylor expansion gives
\[
    \Delta_\eta(\xx, \uu) = \eta \Jac \uu + \frac{\eta^2}{2} \mathbf{H}[\uu, \uu] + O(\eta^3) \, ,
\]
where $\mathbf{H}$ is the second differential of $\Phi$ at $\xx$, a symmetric bilinear map into $\IZ$.
Taking the squared norm and dividing by $\eta^2$ yields
\[
    \norm{\Delta_\eta(\xx, \uu)}^2 / \eta^2 = \norm{\Jac \uu}^2 + \eta \inner{\Jac \uu}{\mathbf{H}[\uu, \uu]} + O(\eta^2) \, .
\]
The map $\uu \mapsto \Jac \uu$ is linear (odd), and $\mathbf{H}[\uu, \uu]$ is a quadratic (even) function of $\uu$, so $\uu \mapsto \inner{\Jac \uu}{\mathbf{H}[\uu, \uu]}$ is odd.
The expectation of an odd function under a law invariant under $\uu \mapsto -\uu$ is zero.
The uniform and singular laws of Section~\ref{sec:laws} have this symmetry, so the term linear in $\eta$ vanishes in expectation and $r_\eta(\xx) = 1 + O(\eta^2)$.
A linear term would bias the reading of the departure from $r_\eta(\xx) = 1$ toward smaller $\eta$.
The simplest such law assigns equal mass to the two directions $\pm\uu$. The odd term cancels between them.
For a single direction taken with one sign only, as for DeepFool in Section~\ref{sec:odin-deepfool}, it does not cancel, and $\norm{\Delta_\eta(\xx, \uu)}^2 / \eta^2$ departs from $\norm{\Jac \uu}^2$ linearly in $\eta$ in general.

For piecewise-affine encoders such as ReLU networks, a Taylor expansion is unnecessary.
The encoder is affine within the activation region containing $\xx$, so $\Delta_\eta(\xx, \uu) = \eta \Jac \uu$ holds exactly along every direction until the perturbation crosses the region boundary.

\paragraph{Sampling.}
For a finite set of directions $U \subset \Sph^{\di - 1}$ drawn from $\mu$, our estimate of the mean squared length of the response per unit scale is
\[
    \frac{1}{|U|} \sum_{\uu \in U} \norm{\Delta_\eta(\xx, \uu)}^2 / \eta^2 \, ,
\]
and our estimate of the mean cosine is the average of the cosines over the directions in $U$ for which we compute the linear prediction $\Jac \uu$.
The mean squared length of the linear prediction, $m_0(\xx)$, is not estimated. Under the uniform law it is computed as $\norm{\Jac}_F^2 / \di$ from the full Jacobian.
For independent draws from a sign-symmetric law, the empirical average of the odd term in the expansion
\[
    \frac{1}{|U|} \sum_{\uu \in U} \inner{\Jac \uu}{\mathbf{H}[\uu, \uu]}
\]
is an average of $|U|$ independent, zero-mean random variables.
It vanishes only in expectation, and its typical size for a single draw is of order $|U|^{-1/2}$.
The estimated size ratio thus retains a term linear in $\eta$.
We sample each direction $\uu \in U$ together with its antipode $-\uu$, so that $U = -U$.
The sum of any odd function over $U$ vanishes exactly by pairwise cancellation, rather than only in expectation.
The estimate then departs from its small-$\eta$ level at order $\eta^2$, matching the earlier derivation.

Under the uniform law, we draw 1024 Gaussian vectors using a fixed seed for each image, shared across all scales and encoders, normalize them to unit length and add their antipodes to form $U$.
The linear prediction $\Jac \uu$ is computed for the first 256 original directions, and the mean cosine is averaged over these without their antipodes, since the cosine has no linear term to cancel.
For the singular laws $\mu_{(k)}$, $U = \{ \vv_k, -\vv_k \}$ exhausts the support of $\mu_{(k)}$, the linear prediction is computed at both directions, and the estimators are exact.

\paragraph{Jacobian and singular vectors.}
For each image, we form the full Jacobian $\Jac$ in single precision from $\dz$ vector-Jacobian products, one per coordinate of the embedding.
The singular values and vectors come from the eigendecomposition of the $\dz \times \dz$ matrix $\Jac \Jac^\top$ in double precision. Its eigenvalues are $\lambda_k$, and the eigenvectors $\mathbf{e}_k$ give $\vv_k = \Jac^\top \mathbf{e}_k / \sqrt{\lambda_k}$.
We treat $\lambda_k$ as numerically positive when $\lambda_k \ge 10^{-6} \lambda_1$.
The Jacobian of ResNet-50 at ImageNet resolution has $2048 \times 150{,}528$ entries and occupies about $1.2$~GB in single precision.
On an NVIDIA H100, computing the Jacobians and their eigendecompositions for the 1000 images takes about $0.2$ to $1.7$ GPU-hours per encoder. The full measurement takes about $9$ to $18$ GPU-hours per encoder.

\paragraph{Relation to the Frobenius norm.}
Our measurement starts from an approximation that motivates the sensitivity measure of~\cite{novak2018sensitivity}.
For a small Gaussian input perturbation $\boldsymbol{\delta} \sim \mathcal{N}(\mathbf{0}, s^2 \mathbf{I})$, the cited work approximates the expected squared change in the output using the linearized network:
\[
    \E_{\boldsymbol{\delta}} \norm{\Phi(\xx + \boldsymbol{\delta}) - \Phi(\xx)}^2 \approx \E_{\boldsymbol{\delta}} \norm{\Jac \boldsymbol{\delta}}^2 = s^2 \norm{\Jac}_F^2 \, .
\]
The cited work uses the right-hand side as a statistic of the network at $\xx$, without examining the range of scales over which the approximation holds.
Replacing the Gaussian law by the uniform law, $\boldsymbol{\delta} = \eta \uu$, gives $\E_{\uu \sim \mu} \norm{\Delta_\eta(\xx, \uu)}^2$ on the left-hand side and $\eta^2 m_0(\xx)$ on the right.
Their ratio is the size ratio $r_\eta(\xx)$, which measures this approximation at finite scales.
The right-hand side of the Gaussian approximation is also the Jacobian penalty of~\cite{hoffman2019robust}.

\subsection{Numerical floor and excluded scales}
\label{app:numerical-floor}

\paragraph{Excluded scales.}
The encoders are evaluated in single precision, and at the smallest scales the response $\Delta_\eta(\xx, \uu)$ approaches floating-point resolution.
The rounding error in $\Delta_\eta(\xx, \uu)$ does not shrink with $\eta$, so after division by $\eta^2$ it contributes a term that grows as $1/\eta^2$.
For each image, we therefore fit the mean squared length of the response per unit scale at the three smallest scales to $\alpha + \beta / \eta^2$.
For a given image, we exclude any scale at which $\beta / \eta^2$ accounts for at least 5\% of the measured value, from both the $r_\eta(\xx)$ and $c_\eta(\xx)$ profiles.
Under the uniform law, this rule excludes 0.056\% of the measured scales across all images and the 27 encoders of Table~\ref{tab:radii_uniform}.
Only the smallest scales are affected, with none excluded beyond the fourth scale (0.037 gray levels). For 14 encoders, all scales are retained.
The same rule applies to each range of spatial frequency.

Under the singular laws, we compare the response with a noise reference instead.
The reference is the median squared length of the response at the smallest scale along 64 kernel directions (Appendix~\ref{app:kernel}), where the linear prediction is zero.
We retain a rank at a given scale only if the squared length of the response, averaged over $\pm\vv_k$, exceeds three times this reference.
For the measurements of Section~\ref{sec:odin-deepfool}, no scale is excluded.

\paragraph{Departure scale.}
We compute the departure scale $\eta^*_\epsilon(\xx)$ from the 17 measured scales by interpolation.
For each of $r_\eta(\xx)$ and $c_\eta(\xx)$, we find the first scale outside its tolerance and the preceding scale. Between these two scales, we interpolate $\log r_\eta(\xx)$ and $c_\eta(\xx)$, respectively, linearly in $\log \eta$.
The departure scale is the smaller of the two scales obtained in this way.
If an image has left the tolerance already at its first scale, we set $\eta^*_\epsilon(\xx)$ to that scale, which is then an upper bound.
If it stays within the tolerance at every scale, we set $\eta^*_\epsilon(\xx) = +\infty$.
Both cases enter the reported medians and quartiles as values.

\subsection{Bootstrap precision}
\label{app:bootstrap-precision}

The departure scale of an image is computed from a finite sample of directions, so we measure how much it depends on that sample.
For each image, we draw 2000 bootstrap samples by resampling its 1024 antipodal pairs with replacement.
Resampling pairs rather than individual directions preserves the sign symmetry of every sample (Appendix~\ref{app:sampling}).
For each resample we recompute $r_\eta(\xx)$, $c_\eta(\xx)$ and $\eta^*_\epsilon(\xx)$, keeping $m_0(\xx)$ and the set of excluded scales (Appendix~\ref{app:numerical-floor}) fixed.
Table~\ref{tab:precision} reports the bootstrap intervals and standard errors together with measures of variation across individual directions.
We did not run the bootstrap for the ResNet-50 encoders trained adversarially with $\ell_2$ budgets (Fig.~\ref{fig:training}, right; Appendix~\ref{app:deepfool-odin}).

\begin{table*}[t]
    \centering
    \caption{
        Bootstrap precision of the departure scale $\eta^*$ and variation across uniformly random directions for the 1000 ImageNet validation images.
        Each image's 2048 directions form 1024 antipodal pairs, resampled 2000 times.
        The $\eta^*$ columns report the multiplicative width $p_{97.5}/p_{2.5}$ of the 95\% bootstrap intervals for $\eta^*_{10\%}$ and $\eta^*_{50\%}$. They also give the share of images whose interval lies within a factor of $10^{1/4}$ of the estimate (the ratio between consecutive measured scales).
        The SE columns report the bootstrap standard errors of $r_\eta$ ($\times 10^{-3}$) and $c_\eta$ ($\times 10^{-4}$) at the measured scale nearest to each image's $\eta^*_{10\%}$.
        At the same scale, the spread columns report the quartile ratio $Q_3/Q_1$ of the per-direction size ratio and the interquartile range of the per-direction cosine ($\times 10^{-3}$). Both $c_\eta$ and the per-direction cosines use the first 256 directions without their antipodes.
        Cells report medians over images, with interquartile ranges.
        For most images, the bootstrap interval for the departure scale lies within a factor of $10^{1/4}$ of the estimate. Individual directions at that scale differ considerably.
    }
    \input{tables/precision}
    \label{tab:precision}
\end{table*}

\subsection{Validation on synthetic encoders}
\label{app:validation}

We validate the measurement on three synthetic encoders whose $r_\eta$ and $c_\eta$ can be predicted from their weights.
Each maps a flattened image to $\dz = 512$ outputs using Gaussian weights drawn once.

The \textit{linear} encoder is $\Phi(\xx) = A \xx$, with the entries of $A$ drawn from $\mathcal{N}(0, 1)$.
Its response equals its linear prediction at every scale, so it never leaves its linear regime.
The \textit{random Fourier features} encoder is
\[
    \Phi(\xx) = \sqrt{2 / \dz} \, \cos(W \xx + \boldsymbol{b}) \, ,
\]
where the cosine is taken componentwise, the entries of $W$ are drawn from $\mathcal{N}(0, \sigma^2)$, and those of $\boldsymbol{b}$ are drawn uniformly from $[0, 2\pi)$.
Its response is sublinear.
The \textit{quadratic} encoder is constructed separately for each image $\xx$ as
\[
    \Phi_{\xx}(\boldsymbol{y}) = A \boldsymbol{y} + \tfrac{1}{2} \, P (\boldsymbol{y} - \xx) \odot Q (\boldsymbol{y} - \xx) \, ,
\]
where $\odot$ denotes the componentwise product and the entries of $P$ and $Q$ are drawn from $\mathcal{N}(0, \sigma^2)$.
Its Jacobian at $\xx$ is $A$, and its response is superlinear.
For the random Fourier features and quadratic encoders, we set $\sigma = 0.655$ and $\sigma = 0.795$, respectively, so that both departure scales $\eta^*_{10\%}$ lie near the middle of the measured range.
The two nonlinear encoders test superlinear and sublinear departures as well as both criteria in the definition of the departure scale.

After averaging over the weights, we obtain the size ratio $r_\eta = (1 - e^{-t}) / t$ with $t = \eta^2 \sigma^2 / 2$ for the random Fourier features encoder, and $r_\eta = 1 + \eta^2 \sigma^4 / 4$ for the quadratic encoder.
Our predictions are computed in the same way for the actual weights drawn and each image, which removes the variation between draws.
The mean cosine is an expectation of a ratio and has no such form, so we approximate it by an expansion with an error of order $\dz^{-2}$.
The predicted departure scale is obtained from these values at the 17 scales using the rule of Appendix~\ref{app:numerical-floor}, so prediction and measurement use the same interpolation.
We evaluate the three encoders on 100 of the 1000 ImageNet validation images, following Appendices~\ref{app:sampling} and~\ref{app:numerical-floor}, except that the mean cosine is averaged over the 256 directions and their antipodes.

Table~\ref{tab:validation} compares predicted and measured departure scales.
The linear encoder shows no departure in any image.
For the two nonlinear encoders, the measured departure scale is within 2\% of the prediction for every image, while consecutive measured scales differ by a factor of $10^{1/4} \approx 1.78$.
For all three encoders, the difference between the measured and predicted $r_\eta$, expressed in units of its bootstrap standard error (Appendix~\ref{app:bootstrap-precision}, here with 1000 resamples), has a mean close to 0 and a standard deviation close to 1.
These results are consistent with an unbiased estimate of $r_\eta$ and a well-calibrated bootstrap standard error.

\begin{table*}[t]
    \centering
    \caption{
        Predicted and measured departure scales of the three synthetic encoders (Appendix~\ref{app:validation}) for 100 ImageNet validation images.
        Departure scales are medians over images; their ratio is taken per image, with its interquartile range in brackets.
        The last column gives the deviation of the measured $r_\eta$ from its prediction in bootstrap standard errors, as mean $\pm$ standard deviation over images and scales.
        The measured departure scales agree with the predicted ones to within 0.1\% in the median, and $r_\eta$ agrees within its standard error.
    }
    \input{tables/validation}
    \label{tab:validation}
\end{table*}

\subsection{Details on encoders}
\label{app:encoders}

Table~\ref{tab:models} lists the encoders we consider and specifies the embedding $\Phi(\xx)$ for each.
For CLIP and SigLIP~2, $\Phi(\xx)$ is the image embedding returned by the model, without unit-length normalization.
For most encoders we use publicly available checkpoints.
We train four groups of models ourselves.

\paragraph{ConvNeXt-T.}
Figure~\ref{fig:training} (left) tracks a ConvNeXt-T over $128$ epochs of ImageNet training (Section~\ref{sec:effect-of-training}).
It is not the ConvNeXt-T checkpoint used in the other experiments.
We train the timm~\cite{rw2019timm} ConvNeXt-T from random initialization on the ImageNet-1k training set, using AdamW with a peak learning rate of $4 \times 10^{-3}$, weight decay of $0.05$ and a batch size of $4096$.
The learning rate is warmed up linearly over the first $2662$ updates and then follows a cosine schedule, for $39{,}936$ updates in total.
We use random resized crops, RandAugment, random erasing, mixup and CutMix, with label smoothing of $0.1$ and stochastic depth of $0.1$.
The final model reaches $80.4\%$ top-1 accuracy on the ImageNet validation set.
Figure~\ref{fig:training} shows the initial weights, $12$ logarithmically spaced checkpoints up to update $6000$, and the checkpoints at the ends of epochs $32$, $64$, $96$ and $128$.
We evaluate these $17$ checkpoints on the first $256$ of the 1000 images.

\paragraph{CIFAR-10 ResNets.}
Figure~\ref{fig:depth-family-rn} compares the departure scale at initialization and after training in CIFAR-10 ResNets of depth $8$ to $218$ (Section~\ref{sec:effect-of-training}).
The CIFAR-10 ResNets~\cite{he2016deep} have depths $8$, $14$, $20$, $32$, $44$, $56$, $110$ and $218$, with three training seeds per depth.
We train on $45{,}000$ training images and hold out the remaining $5000$ ($500$ per class).
Training uses SGD with momentum $0.9$, weight decay $10^{-4}$ and batch size $128$ for $64{,}000$ updates.
The learning rate is $0.1$, divided by $10$ after $32{,}000$ and $48{,}000$ updates. At every depth, the first $400$ updates use a learning rate of $0.01$.
We use random crops with $4$ pixels of padding and horizontal flips.
After training, the mean test accuracy over seeds rises from $87.6\%$ at depth $8$ to $93.1\%$ at depths $110$ and $218$.
We measure the initialization and the final weights on $1000$ of the held-out images, $100$ per class.
At initialization, we set the running mean and variance of every batch-norm layer to the exact statistics of its input over the $45{,}000$ training images, layer by layer in forward order, with all weights left at their initial values.

\paragraph{CIFAR-10 ResNet-18.}
The memorization experiment of Section~\ref{sec:effect-of-training} uses ResNet-18 on CIFAR-10 with ReLU or GELU activations, each trained with true labels and with shuffled labels (Fig.~\ref{fig:training}, middle).
The network is the torchvision~\cite{torchvision2016} ResNet-18 with a $3 \times 3$ stem of stride $1$ and no max pooling. In the GELU version, every ReLU is replaced with GELU.
Each of the four networks is trained on all $50{,}000$ training images for $200$ epochs, using SGD with momentum $0.9$, a learning rate of $0.01$ with a cosine schedule, and batch size $128$. There is no weight decay or augmentation.
The same permutation of the true training labels is used for both activation functions.
Both networks trained on shuffled labels achieve $100\%$ training accuracy on those labels and $10.5\%$ test accuracy.
When trained on true labels, the networks achieve test accuracies of $85.4\%$ with ReLU and $85.3\%$ with GELU.
We evaluate the trained networks on $1000$ images sampled at random from the test set.

\paragraph{Fine-tuning.}
Appendix~\ref{app:effect-of-training} repeats the memorization experiment on the three supervised ImageNet encoders by fine-tuning them on the 1000 measured images, with true labels and with shuffled labels (Fig.~\ref{fig:memorization}, Table~\ref{tab:radii_mem}).
Starting from the supervised checkpoints of Table~\ref{tab:models}, we fine-tune all weights using the 1000 images, preprocessed as for the measurements, without augmentation.
Training uses AdamW with a constant learning rate of $10^{-4}$, no weight decay and batch size $64$. The networks remain in evaluation mode, so the batch-norm statistics stay fixed.
The shuffled labels are obtained by permuting the true labels of the 1000 images.
With shuffled labels, we train until the network predicts the assigned label for every image, which takes $17$ epochs for ResNet-50, $11$ for ConvNeXt-T and $8$ for ViT-S/16.
With the true labels, training runs for the same number of updates.

\begin{table*}[t]
    \centering
    \caption{
        Encoders used in our experiments, with their weight sources, training regimes, embedding definitions $\Phi(\xx)$ and embedding dimensions $\dz$.
        For classifiers, $\Phi(\xx)$ is the input to the final linear layer.
        Rows marked $\varepsilon = 0$ denote the models trained without adversarial examples, which serve as controls for the adversarially trained models.
    }
    \input{tables/models}
    \label{tab:models}
\end{table*}

%% file: tables/precision.tex
\begingroup
\scriptsize
\setlength{\tabcolsep}{2.5pt}
\renewcommand{\arraystretch}{1.32}
\begin{tabular}{@{}c@{\hspace{6pt}}l@{\hspace{8pt}}r@{\,}lr@{\,}lr@{\hspace{12pt}}r@{\,}lr@{\,}l@{\hspace{12pt}}r@{\,}lr@{\,}l@{}}
  \toprule
    & & \multicolumn{5}{c}{$\eta^\ast$: 95\% CI} & \multicolumn{4}{c}{SE at $\eta^\ast_{10\%}$} & \multicolumn{4}{c}{Spread at $\eta^\ast_{10\%}$} \\
  \cmidrule(lr){3-7}\cmidrule(lr){8-11}\cmidrule(l){12-15}
  & \textbf{Encoder} & \multicolumn{2}{c}{10\%} & \multicolumn{2}{c}{50\%} & within one scale (\%) & \multicolumn{2}{c}{$r_\eta$} & \multicolumn{2}{c}{$c_\eta$} & \multicolumn{2}{c}{$q/q_0$} & \multicolumn{2}{c}{cosine} \\
  \midrule
  \multirow{3}{*}{\rotatebox[origin=c]{90}{\textbf{\textit{Supervised}}}} & ResNet-50 & 1.07 & [1.05, 1.09] & 1.08 & [1.07, 1.10] & 99 & 7.0 & [5.9, 8.7] & 18.3 & [15.0, 23.0] & 1.20 & [1.17, 1.25] & 37.0 & [30.6, 45.9] \\
   & ConvNeXt-T & 1.14 & [1.11, 1.18] & 1.02 & [1.02, 1.02] & 100 & 7.3 & [6.4, 8.5] & 8.3 & [6.1, 11.4] & 1.16 & [1.13, 1.20] & 17.6 & [12.8, 24.7] \\
   & ViT-S/16 & 1.32 & [1.18, 1.76] & 1.10 & [1.04, 1.13] & 84 & 15.9 & [11.6, 24.0] & 27.6 & [16.2, 44.1] & 1.32 & [1.24, 1.43] & 37.9 & [22.5, 59.1] \\
  \addlinespace[2pt]\arrayrulecolor{black!60}\cmidrule{1-15}\arrayrulecolor{black}\addlinespace[2pt]
  \multirow{3}{*}{\rotatebox[origin=c]{90}{\textbf{\textit{Untrained}}}} & ResNet-50 & 1.03 & [1.03, 1.04] & 1.01 & [1.01, 1.01] & 100 & 3.4 & [3.3, 3.5] & 11.3 & [9.0, 14.4] & 1.18 & [1.16, 1.21] & 24.1 & [19.3, 30.5] \\
   & ConvNeXt-T & 1.05 & [1.04, 1.07] & --- &  & 99 & 2.0 & [1.9, 2.1] & 2.2 & [1.6, 3.4] & 1.04 & [1.04, 1.05] & 4.6 & [3.3, 7.2] \\
   & ViT-S/16 & 1.10 & [1.08, 1.13] & 1.01 & [1.01, 1.02] & 100 & 5.6 & [5.3, 5.8] & 6.3 & [4.9, 8.6] & 1.10 & [1.08, 1.12] & 13.5 & [10.3, 18.1] \\
  \addlinespace[2pt]\arrayrulecolor{black!60}\cmidrule{1-15}\arrayrulecolor{black}\addlinespace[2pt]
  \multirow{12}{*}{\rotatebox[origin=c]{90}{\textbf{\textit{Adversarial}}}} & ResNet-50 & \multicolumn{13}{@{}l@{}}{{\color{black!25}\xleaders\hrule height 0.62ex depth -0.55ex\hskip 0pt plus 1filll}} \\
   & \quad $\varepsilon = 0$ & 1.06 & [1.04, 1.07] & 1.07 & [1.02, 1.08] & 100 & 7.0 & [6.3, 7.8] & 17.6 & [14.6, 22.0] & 1.20 & [1.17, 1.24] & 36.6 & [30.2, 45.0] \\
   & \quad $\ell_\infty$ 1/255 & 1.17 & [1.13, 1.22] & 1.02 & [1.01, 1.02] & 99 & 8.5 & [7.6, 9.8] & 9.9 & [7.8, 12.8] & 1.17 & [1.14, 1.20] & 21.1 & [16.4, 26.7] \\
   & \quad $\ell_\infty$ 2/255 & 1.18 & [1.14, 1.24] & 1.02 & [1.02, 1.02] & 100 & 8.9 & [7.9, 10.1] & 10.8 & [8.5, 13.9] & 1.17 & [1.14, 1.21] & 22.4 & [17.9, 28.5] \\
   & \quad $\ell_\infty$ 4/255 & 1.25 & [1.17, 1.39] & 1.02 & [1.02, 1.03] & 99 & 9.3 & [8.3, 10.8] & 12.0 & [9.5, 14.6] & 1.17 & [1.15, 1.21] & 24.9 & [19.5, 30.3] \\
   & \quad $\ell_\infty$ 8/255 & 1.27 & [1.00, 1.43] & 1.02 & [1.02, 1.05] & 98 & 10.3 & [9.0, 13.0] & 22.4 & [16.0, 56.2] & 1.28 & [1.22, 1.44] & 44.4 & [31.8, 84.7] \\
   & ConvNeXt-T & \multicolumn{13}{@{}l@{}}{{\color{black!25}\xleaders\hrule height 0.62ex depth -0.55ex\hskip 0pt plus 1filll}} \\
   & \quad $\varepsilon = 0$ & 1.17 & [1.12, 1.24] & 1.02 & [1.02, 1.04] & 99 & 8.8 & [7.4, 12.7] & 11.3 & [8.0, 16.9] & 1.20 & [1.15, 1.29] & 23.6 & [16.8, 35.0] \\
   & \quad $\ell_\infty$ 4/255 & 1.21 & [1.16, 1.27] & 1.02 & [1.02, 1.03] & 100 & 10.5 & [9.2, 12.3] & 9.4 & [7.1, 12.7] & 1.16 & [1.13, 1.20] & 20.1 & [14.9, 26.6] \\
   & ViT-S/16 & \multicolumn{13}{@{}l@{}}{{\color{black!25}\xleaders\hrule height 0.62ex depth -0.55ex\hskip 0pt plus 1filll}} \\
   & \quad $\varepsilon = 0$ & 1.20 & [1.11, 1.41] & 1.12 & [1.09, 1.17] & 96 & 14.1 & [11.4, 17.6] & 33.9 & [23.7, 45.9] & 1.38 & [1.28, 1.52] & 53.5 & [37.8, 71.3] \\
   & \quad $\ell_\infty$ 4/255 & 1.23 & [1.18, 1.29] & 1.02 & [1.02, 1.03] & 99 & 11.3 & [10.0, 13.3] & 10.2 & [7.9, 13.3] & 1.18 & [1.14, 1.22] & 21.4 & [16.2, 28.1] \\
  \addlinespace[2pt]\arrayrulecolor{black!60}\cmidrule{1-15}\arrayrulecolor{black}\addlinespace[2pt]
  \multirow{9}{*}{\rotatebox[origin=c]{90}{\textbf{\textit{Memorization}}}} & ResNet-50 & \multicolumn{13}{@{}l@{}}{{\color{black!25}\xleaders\hrule height 0.62ex depth -0.55ex\hskip 0pt plus 1filll}} \\
   & \quad true labels & 1.07 & [1.06, 1.09] & 1.08 & [1.07, 1.10] & 99 & 7.7 & [6.8, 9.0] & 19.1 & [15.9, 23.2] & 1.21 & [1.18, 1.25] & 38.4 & [31.7, 46.3] \\
   & \quad shuffled labels & 1.07 & [1.05, 1.09] & 1.05 & [1.02, 1.08] & 99 & 8.0 & [7.0, 10.1] & 20.8 & [16.7, 26.7] & 1.25 & [1.21, 1.31] & 42.4 & [34.3, 53.8] \\
   & ConvNeXt-T & \multicolumn{13}{@{}l@{}}{{\color{black!25}\xleaders\hrule height 0.62ex depth -0.55ex\hskip 0pt plus 1filll}} \\
   & \quad true labels & 1.13 & [1.10, 1.16] & 1.02 & [1.01, 1.02] & 100 & 6.4 & [5.8, 7.3] & 7.4 & [5.5, 10.0] & 1.13 & [1.11, 1.16] & 15.9 & [11.6, 21.0] \\
   & \quad shuffled labels & 1.15 & [1.10, 1.20] & 1.05 & [1.02, 1.07] & 100 & 8.2 & [6.9, 10.1] & 11.7 & [8.5, 16.0] & 1.18 & [1.15, 1.24] & 24.8 & [17.8, 33.1] \\
   & ViT-S/16 & \multicolumn{13}{@{}l@{}}{{\color{black!25}\xleaders\hrule height 0.62ex depth -0.55ex\hskip 0pt plus 1filll}} \\
   & \quad true labels & 1.32 & [1.17, 1.78] & 1.11 & [1.08, 1.15] & 85 & 15.8 & [11.1, 23.9] & 32.1 & [19.5, 48.6] & 1.28 & [1.20, 1.38] & 39.2 & [22.0, 65.0] \\
   & \quad shuffled labels & 1.15 & [1.10, 1.21] & 1.04 & [1.02, 1.08] & 97 & 7.8 & [6.9, 9.8] & 19.5 & [12.7, 29.8] & 1.23 & [1.18, 1.30] & 31.7 & [21.2, 44.2] \\
  \addlinespace[2pt]\arrayrulecolor{black!60}\cmidrule{1-15}\arrayrulecolor{black}\addlinespace[2pt]
  \multirow{3}{*}{\rotatebox[origin=c]{90}{\textbf{\textit{SSL}}}} & DINOv2 ViT-S/14 & 1.15 & [1.11, 1.19] & 1.02 & [1.02, 1.06] & 100 & 7.6 & [6.6, 9.0] & 9.4 & [6.9, 12.7] & 1.15 & [1.13, 1.19] & 19.8 & [14.8, 26.8] \\
   & DINOv2 ViT-S/14 (head input) & 1.13 & [1.10, 1.17] & 1.02 & [1.02, 1.03] & 100 & 7.0 & [6.1, 8.2] & 8.2 & [6.1, 11.1] & 1.15 & [1.12, 1.19] & 17.4 & [12.9, 23.8] \\
   & ConvNeXt-V2-T & 1.17 & [1.12, 1.23] & 1.06 & [1.03, 1.10] & 99 & 8.6 & [7.5, 10.8] & 12.6 & [9.2, 17.8] & 1.20 & [1.15, 1.26] & 26.5 & [19.4, 36.6] \\
  \addlinespace[2pt]\arrayrulecolor{black!60}\cmidrule{1-15}\arrayrulecolor{black}\addlinespace[2pt]
  \multirow{2}{*}{\rotatebox[origin=c]{90}{\textbf{\textit{VLM}}}} & CLIP ViT-B/16 & 1.31 & [1.21, 1.45] & 1.04 & [1.03, 1.06] & 96 & 15.5 & [12.5, 19.3] & 22.1 & [14.2, 33.0] & 1.40 & [1.31, 1.53] & 40.7 & [27.3, 60.2] \\
   & SigLIP 2 ViT-B/16 & 1.18 & [1.08, 1.36] & 1.07 & [1.04, 1.10] & 95 & 14.5 & [11.5, 18.8] & 24.9 & [17.6, 35.5] & 1.36 & [1.26, 1.50] & 45.1 & [33.1, 61.9] \\
  \bottomrule
\end{tabular}
\endgroup

%% file: tables/validation.tex
\begingroup
\small
\setlength{\tabcolsep}{2.5pt}
\renewcommand{\arraystretch}{1.32}
\begin{tabular}{@{}l@{\hspace{8pt}}rrr@{\,}l@{\hspace{12pt}}rrr@{\,}l@{\hspace{12pt}}r@{}}
  \toprule
    & \multicolumn{4}{c}{$\eta^\ast_{10\%}$} & \multicolumn{4}{c}{$\eta^\ast_{50\%}$} & \\
  \cmidrule(lr){2-5}\cmidrule(lr){6-9}
  \textbf{Encoder} & predicted & measured & \multicolumn{2}{c}{measured/predicted} & predicted & measured & \multicolumn{2}{c}{measured/predicted} & $z$ of $r_\eta$ \\
  \midrule
  Linear & \multicolumn{8}{c}{no departure in all 100 images} & $-0.04 \pm 1.08$ \\
  Random Fourier features & 0.498 & 0.498 & 1.000 & [0.998, 1.003] & 1.340 & 1.340 & 1.000 & [0.999, 1.001] & $-0.04 \pm 0.99$ \\
  Quadratic & 0.679 & 0.680 & 1.001 & [0.996, 1.004] & 2.080 & 2.081 & 1.000 & [0.999, 1.001] & $-0.05 \pm 1.07$ \\
  \bottomrule
\end{tabular}
\endgroup

%% file: tables/models.tex
\begingroup
\scriptsize
\setlength{\tabcolsep}{2.5pt}
\renewcommand{\arraystretch}{1.32}
\begin{tabular}{@{}c@{\hspace{6pt}}l@{\hspace{8pt}}lllr@{}}
  \toprule
  & \textbf{Model} & Weights & Training & $\Phi$ & $d_Z$ \\
  \midrule
  \multirow{5}{*}{\rotatebox[origin=c]{90}{\textbf{\textit{Supervised}}}} & ResNet-50~\cite{he2016deep} & torchvision, IMAGENET1K\_V2~\cite{torchvision2016} & ImageNet-1k, TorchVision recipe~\cite{vryniotis2021how} & global average pool & 2048 \\
   & ResNet-50 V1~\cite{he2016deep} & torchvision, IMAGENET1K\_V1~\cite{torchvision2016} & ImageNet-1k; in OpenOOD~\cite{yang2022openood,zhang2024openood} & global average pool & 2048 \\
   & ConvNeXt-T~\cite{liu2022convnet} & torchvision, IMAGENET1K\_V1~\cite{torchvision2016} & ImageNet-1k, TorchVision recipe~\cite{vryniotis2021how} & average pool, LayerNorm & 768 \\
   & ConvNeXt-T, over training~\cite{liu2022convnet} & ours, timm convnext\_tiny~\cite{rw2019timm} & ImageNet-1k, 128 epochs & average pool, LayerNorm & 768 \\
   & ViT-S/16~\cite{dosovitskiy2021image,touvron2021training} & timm, vit\_small\_patch16\_224.augreg\_in1k~\cite{rw2019timm} & ImageNet-1k, AugReg~\cite{steiner2022how} & class token, final norm & 384 \\
  \addlinespace[2pt]\arrayrulecolor{black!60}\cmidrule{1-6}\arrayrulecolor{black}\addlinespace[2pt]
  \multirow{3}{*}{\rotatebox[origin=c]{90}{\textbf{\textit{Untrained}}}} & ResNet-50~\cite{he2016deep} & torchvision initialization, seed 0~\cite{torchvision2016} & He init. (torchvision default)~\cite{he2015delving} & global average pool & 2048 \\
   & ConvNeXt-T~\cite{liu2022convnet} & torchvision initialization, seed 0~\cite{torchvision2016} & trunc. normal, $\sigma = 0.02$~\cite{torchvision2016} & average pool, LayerNorm & 768 \\
   & ViT-S/16~\cite{dosovitskiy2021image,touvron2021training} & timm initialization, seed 0~\cite{rw2019timm} & trunc. normal, $\sigma = 0.02$~\cite{rw2019timm} & class token, final norm & 384 \\
  \addlinespace[2pt]\arrayrulecolor{black!60}\cmidrule{1-6}\arrayrulecolor{black}\addlinespace[2pt]
  \multirow{20}{*}{\rotatebox[origin=c]{90}{\textbf{\textit{Adversarial}}}} & ResNet-50~\cite{he2016deep} & \multicolumn{4}{@{}l@{}}{{\color{black!25}\xleaders\hrule height 0.62ex depth -0.55ex\hskip 0pt plus 1filll}} \\
   & \quad $\varepsilon = 0$ & authors' release, resnet50\_l2\_eps0~\cite{salman2020adversarially} & ImageNet-1k, standard & global average pool & 2048 \\
   & \quad $\varepsilon = 0$, second run & authors' release, resnet50\_linf\_eps0~\cite{salman2020adversarially} & ImageNet-1k, standard & global average pool & 2048 \\
   & \quad $\ell_\infty$ 1/255 & authors' release, resnet50\_linf\_eps1.0~\cite{salman2020adversarially} & ImageNet-1k, adversarial & global average pool & 2048 \\
   & \quad $\ell_\infty$ 2/255 & authors' release, resnet50\_linf\_eps2.0~\cite{salman2020adversarially} & ImageNet-1k, adversarial & global average pool & 2048 \\
   & \quad $\ell_\infty$ 4/255 & authors' release, resnet50\_linf\_eps4.0~\cite{salman2020adversarially} & ImageNet-1k, adversarial & global average pool & 2048 \\
   & \quad $\ell_\infty$ 8/255 & authors' release, resnet50\_linf\_eps8.0~\cite{salman2020adversarially} & ImageNet-1k, adversarial & global average pool & 2048 \\
   & \quad $\ell_2$ 0.03 & authors' release, resnet50\_l2\_eps0.03~\cite{salman2020adversarially} & ImageNet-1k, adversarial & global average pool & 2048 \\
   & \quad $\ell_2$ 0.1 & authors' release, resnet50\_l2\_eps0.1~\cite{salman2020adversarially} & ImageNet-1k, adversarial & global average pool & 2048 \\
   & \quad $\ell_2$ 0.25 & authors' release, resnet50\_l2\_eps0.25~\cite{salman2020adversarially} & ImageNet-1k, adversarial & global average pool & 2048 \\
   & \quad $\ell_2$ 0.5 & authors' release, resnet50\_l2\_eps0.5~\cite{salman2020adversarially} & ImageNet-1k, adversarial & global average pool & 2048 \\
   & \quad $\ell_2$ 1 & authors' release, resnet50\_l2\_eps1~\cite{salman2020adversarially} & ImageNet-1k, adversarial & global average pool & 2048 \\
   & \quad $\ell_2$ 3 & authors' release, resnet50\_l2\_eps3~\cite{salman2020adversarially} & ImageNet-1k, adversarial & global average pool & 2048 \\
   & \quad $\ell_2$ 5 & authors' release, resnet50\_l2\_eps5~\cite{salman2020adversarially} & ImageNet-1k, adversarial & global average pool & 2048 \\
   & ConvNeXt-T~\cite{liu2022convnet} & \multicolumn{4}{@{}l@{}}{{\color{black!25}\xleaders\hrule height 0.62ex depth -0.55ex\hskip 0pt plus 1filll}} \\
   & \quad $\varepsilon = 0$ & timm, convnext\_tiny.fb\_in1k~\cite{rw2019timm} & ImageNet-1k~\cite{liu2022convnet} & average pool, LayerNorm & 768 \\
   & \quad $\ell_\infty$ 4/255 & authors' release, weights\_convnext\_t~\cite{singh2023revisiting} & from $\varepsilon = 0$, adversarial & average pool, LayerNorm & 768 \\
   & ViT-S/16~\cite{dosovitskiy2021image,touvron2021training} & \multicolumn{4}{@{}l@{}}{{\color{black!25}\xleaders\hrule height 0.62ex depth -0.55ex\hskip 0pt plus 1filll}} \\
   & \quad $\varepsilon = 0$ & timm, deit\_small\_patch16\_224.fb\_in1k~\cite{rw2019timm} & ImageNet-1k, DeiT~\cite{touvron2021training} & class token, final norm & 384 \\
   & \quad $\ell_\infty$ 4/255 & authors' release, weights\_vit\_s~\cite{singh2023revisiting} & from $\varepsilon = 0$, adversarial & class token, final norm & 384 \\
  \addlinespace[2pt]\arrayrulecolor{black!60}\cmidrule{1-6}\arrayrulecolor{black}\addlinespace[2pt]
  \multirow{9}{*}{\rotatebox[origin=c]{90}{\textbf{\textit{Memorization}}}} & ResNet-50~\cite{he2016deep} & \multicolumn{4}{@{}l@{}}{{\color{black!25}\xleaders\hrule height 0.62ex depth -0.55ex\hskip 0pt plus 1filll}} \\
   & \quad true labels & ours, from Supervised ResNet-50 & the 1000 images, true labels & global average pool & 2048 \\
   & \quad shuffled labels & ours, from Supervised ResNet-50 & the 1000 images, shuffled labels~\cite{zhang2017understanding} & global average pool & 2048 \\
   & ConvNeXt-T~\cite{liu2022convnet} & \multicolumn{4}{@{}l@{}}{{\color{black!25}\xleaders\hrule height 0.62ex depth -0.55ex\hskip 0pt plus 1filll}} \\
   & \quad true labels & ours, from Supervised ConvNeXt-T & the 1000 images, true labels & average pool, LayerNorm & 768 \\
   & \quad shuffled labels & ours, from Supervised ConvNeXt-T & the 1000 images, shuffled labels~\cite{zhang2017understanding} & average pool, LayerNorm & 768 \\
   & ViT-S/16~\cite{dosovitskiy2021image,touvron2021training} & \multicolumn{4}{@{}l@{}}{{\color{black!25}\xleaders\hrule height 0.62ex depth -0.55ex\hskip 0pt plus 1filll}} \\
   & \quad true labels & ours, from Supervised ViT-S/16 & the 1000 images, true labels & class token, final norm & 384 \\
   & \quad shuffled labels & ours, from Supervised ViT-S/16 & the 1000 images, shuffled labels~\cite{zhang2017understanding} & class token, final norm & 384 \\
  \addlinespace[2pt]\arrayrulecolor{black!60}\cmidrule{1-6}\arrayrulecolor{black}\addlinespace[2pt]
  \multirow{3}{*}{\rotatebox[origin=c]{90}{\textbf{\textit{SSL}}}} & DINOv2 ViT-S/14~\cite{dosovitskiy2021image,touvron2021training} & authors' release, dinov2\_vits14~\cite{oquab2024dinov2} & LVD-142M, distilled from ViT-g/14 & class token, final norm & 384 \\
   & DINOv2 ViT-S/14 (head input)~\cite{dosovitskiy2021image,touvron2021training} & dinov2\_vits14 with its linear4 head~\cite{oquab2024dinov2} & LVD-142M, distilled from ViT-g/14 & 4 class tokens, mean patch token & 1920 \\
   & ConvNeXt-V2-T~\cite{woo2023convnext} & timm, convnextv2\_tiny.fcmae~\cite{rw2019timm} & ImageNet-1k images, FCMAE~\cite{woo2023convnext} & average pool, no LayerNorm & 768 \\
  \addlinespace[2pt]\arrayrulecolor{black!60}\cmidrule{1-6}\arrayrulecolor{black}\addlinespace[2pt]
  \multirow{2}{*}{\rotatebox[origin=c]{90}{\textbf{\textit{VLM}}}} & CLIP ViT-B/16~\cite{dosovitskiy2021image} & openai/clip-vit-base-patch16~\cite{wolf2020transformers} & WIT, CLIP~\cite{radford2021learning} & image embedding & 512 \\
   & SigLIP 2 ViT-B/16~\cite{dosovitskiy2021image} & google/siglip2-base-patch16-224~\cite{wolf2020transformers} & WebLI, SigLIP 2~\cite{tschannen2025siglip} & image embedding & 768 \\
  \addlinespace[2pt]\arrayrulecolor{black!60}\cmidrule{1-6}\arrayrulecolor{black}\addlinespace[2pt]
  \multirow{6}{*}{\rotatebox[origin=c]{90}{\textbf{\textit{CIFAR-10}}}} & ResNet, depth 8 to 218~\cite{he2016deep} & \multicolumn{4}{@{}l@{}}{{\color{black!25}\xleaders\hrule height 0.62ex depth -0.55ex\hskip 0pt plus 1filll}} \\
   & \quad trained & ours, seeds 0, 1, 2 & CIFAR-10, recipe of~\cite[Sec.~4.2]{he2016deep} & global average pool & 64 \\
   & \quad initialization & ours, batch norm recalibrated & He init., BN statistics recalibrated~\cite{he2015delving} & global average pool & 64 \\
   & ResNet-18~\cite{he2016deep} & \multicolumn{4}{@{}l@{}}{{\color{black!25}\xleaders\hrule height 0.62ex depth -0.55ex\hskip 0pt plus 1filll}} \\
   & \quad true labels & ours, ReLU and GELU & CIFAR-10, true labels & global average pool & 512 \\
   & \quad shuffled labels & ours, ReLU and GELU & CIFAR-10, shuffled labels~\cite{zhang2017understanding} & global average pool & 512 \\
  \bottomrule
\end{tabular}
\endgroup

%% file: sections/appendix/B.tex
\section{Additional results}

\begin{figure}
    \includegraphics{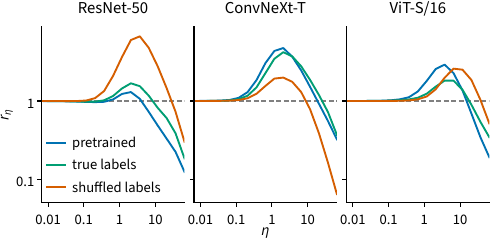}
    \caption{
         Median size ratio $r_\eta$ under the uniform law for the three supervised encoders at their pretrained checkpoints and after fine-tuning on the 1000 measured ImageNet validation images with true or shuffled labels.
    }
    \label{fig:memorization}
\end{figure}

\begin{figure}
    \includegraphics[width=\linewidth]{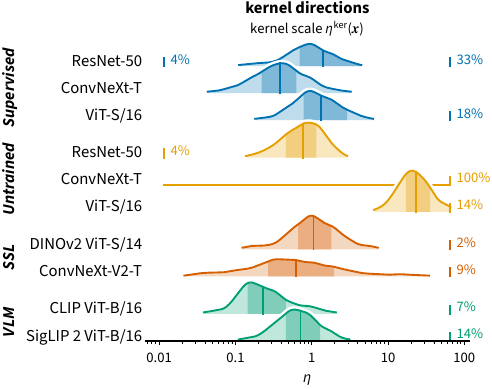}
    \caption{
        Distributions of kernel scales $\eta^{\ker}$ in gray levels for the 1000 ImageNet validation images, with medians and interquartile ranges.
        Percentages on the left give the share of images whose kernel scale lies at or below the smallest measured scale. Those on the right give the share for which the kernel ratio does not reach $1$ over the measured range.
    }
    \label{fig:kernel-scale}
\end{figure}

\subsection{Beyond the linear regime}
\label{app:beyond-linear-regime}
We ask at which singular ranks and in which ranges of spatial frequency the median size ratio rises above its linear prediction.
We describe how we sample the singular directions and the spatial-frequency ranges in Fig.~\ref{fig:singular-freq-heatmaps}, and report the resulting size ratios and departure scales.

\paragraph{Singular ranks.}
For the singular laws $\mu_{(k)}$ of Section~\ref{sec:laws}, we measure the response along the pair $\pm \vv_k$ at ranks $k = 1, \dots, 16$ and at every power of two up to $\dz$.
For each image, we include a rank only if its $\lambda_k$ is numerically positive.
Beyond the transition rank in Fig.~\ref{fig:singular-freq-heatmaps}, the peak of the median size ratio grows with rank, reaching $8.7$ for ConvNeXt-T and $12.5$ for ViT-S/16 at their highest measured ranks.
For ResNet-50, the peak reaches $74$ at rank $1024$, a value measured on $164$ images. In the remaining images, $\lambda_{1024}$ is numerically zero.

\paragraph{Spatial frequency.}
To draw from the uniform law restricted to a range of spatial frequency, we generate a Gaussian image, zero the Fourier coefficients outside the range and normalize the result to unit length.
The subspaces for the four frequency ranges, $1$--$4$, $4$--$16$, $16$--$64$ and above $64$ cycles per image, have dimensions $132$, $2244$, $36168$ and $111981$, respectively. The highest range accounts for about three quarters of the input dimensions.
For each image and range, we draw $64$ such directions with their antipodes and use $m_0(\xx) = \norm{\Jac P}_F^2 / \dim P$, where $P$ is the orthogonal projection onto the corresponding subspace.
In the three higher ranges, the median departure scale $\eta^*_{10\%}$ is no larger than under the uniform law.
In the lowest range, the median is about twice the uniform-law value for ResNet-50 and three times that value for ConvNeXt-T, but is nearly unchanged for ViT-S/16.
The median size ratio exceeds $1$ in every range, but has the smallest peak in the lowest range. There it reaches $1.04$ for ResNet-50, $3.2$ for ConvNeXt-T and $1.35$ for ViT-S/16.

\begin{figure*}
    \includegraphics{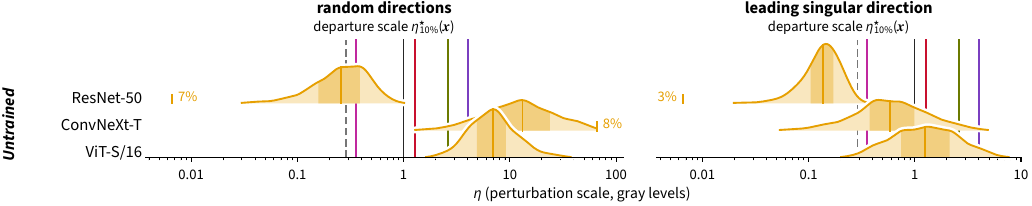}
    \caption{
        Distributions of departure scales $\eta^*_{10\%}$ for the three untrained encoders over the 1000 ImageNet validation images.
        Vertical lines are the reference scales of Fig.~\ref{fig:departure_distributions}.
        Percentages on the left give the share of images with departure scales below the smallest measured scale. Those on the right give the share with no departure over the measured range.
        Along random directions, the linear regimes of untrained ConvNeXt-T and ViT-S/16 extend beyond all reference scales for most images.
    }
    \label{fig:departure-distribution-untrained}
\end{figure*}

\begin{figure*}
    \centering
    \includegraphics[width=0.49\linewidth]{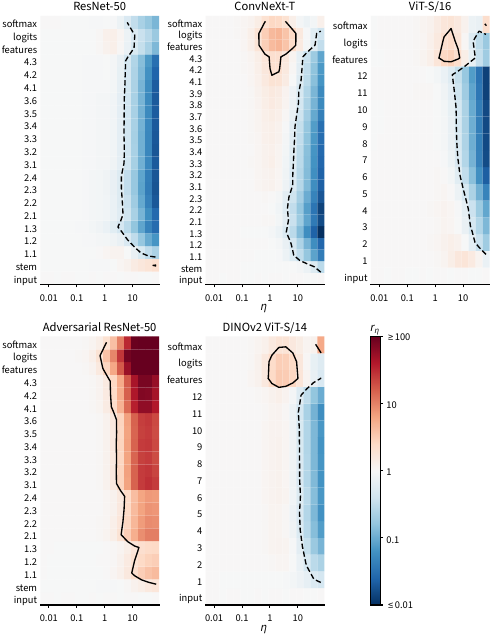}\hfill
    \includegraphics[width=0.49\linewidth]{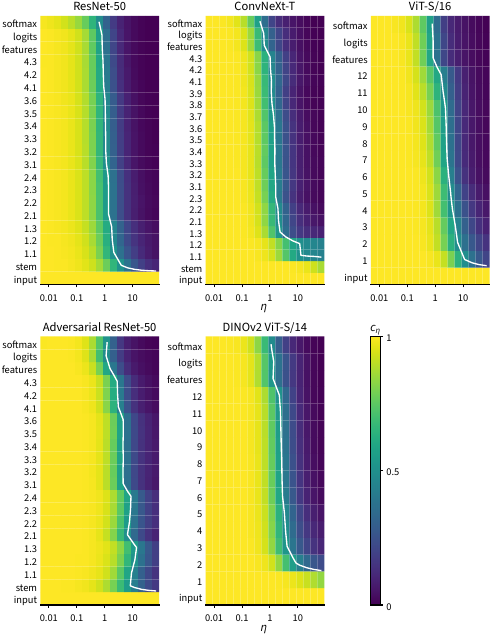}
    \caption{
        Layerwise size ratio $r_\eta$ (left) and mean cosine $c_\eta$ (right) under the uniform law, reported as medians over the 1000 ImageNet validation images.
        The row labeled \emph{features} corresponds to the embedding used in the rest of the paper.
        The adversarially trained ResNet-50 uses an $\ell_2$ budget of $\varepsilon = 3$.
        Colors and lines are as in Fig.~\ref{fig:singular-freq-heatmaps}.
    }
    \label{fig:depth_rc}
\end{figure*}

\subsection{The effect of training}
\label{app:effect-of-training}
We report the departure scales at initialization, repeat the memorization experiment on the supervised ImageNet encoders and examine adversarial training with $\ell_\infty$ budgets.

\paragraph{Initialization.}
The three untrained encoders differ widely in their median departure scales.
Figure~\ref{fig:departure-distribution-untrained} uses the format of Fig.~\ref{fig:departure_distributions} to show the departure scales of the three untrained encoders. Table~\ref{tab:radii_uniform} lists their values next to those of the trained encoders.
At a tolerance of $10\%$, the median departure scales are $0.26$ gray levels for ResNet-50, $13.2$ for ConvNeXt-T and $6.9$ for ViT-S/16.
At a tolerance of $50\%$, the median is $0.83$ for ResNet-50 and $25.7$ for ViT-S/16, while for untrained ConvNeXt-T more than half of the images have no departure within the measured range.

\paragraph{Memorization.}
The memorization experiment of Section~\ref{sec:effect-of-training} uses ResNet-18 trained from scratch on CIFAR-10 with shuffled labels (Fig.~\ref{fig:training}, middle).
Here we fine-tune the three supervised ImageNet encoders on the 1000 images of our measurements, once with shuffled labels and once with the true labels, and repeat the measurements on the same set (Fig.~\ref{fig:memorization}, Table~\ref{tab:radii_mem}).
Fine-tuning on true labels leaves the profiles close to those of the pretrained encoders. The median departure scale under the uniform law increases by a factor of $1.2$ to $1.4$.
If the result of training from scratch with shuffled labels carried over to fine-tuning, the superlinear response would disappear.
With shuffled labels, the three encoders differ.
The median departure scale remains at $0.15$ gray levels for ResNet-50, is about twice as large for ConvNeXt-T (from $0.12$ to $0.23$) and is five times as large for ViT-S/16 (from $0.16$ to $0.78$).
The superlinear response persists in all three. Its peak is higher for ResNet-50, lower for ConvNeXt-T, and shifts to larger scales for ViT-S/16.
Unlike training from scratch with shuffled labels, fine-tuning with them does not remove the superlinear response of the pretrained encoder.

\paragraph{$\ell_\infty$ adversarial training.}
Adversarial training with $\ell_\infty$ budgets can increase the median departure scale without bringing it to one gray level.
Table~\ref{tab:radii_uniform} also reports departure scales for encoders trained adversarially with $\ell_\infty$ budgets.
For ResNet-50~\cite{salman2020adversarially}, the median departure scale increases from $0.23$ gray levels without adversarial training to $0.46$, $0.43$ and $0.38$ at budgets of $1/255$, $2/255$ and $4/255$, respectively, but falls to $0.09$ at $8/255$.
At a budget of $4/255$, the median increases from $0.13$ to $0.42$ for ConvNeXt-T and from $0.23$ to $0.77$ for ViT-S/16.
Up to $4/255$, the linear regime is two to three times as long as without adversarial training, which the $\ell_2$ family does not show.
It remains below one gray level in every case, while these budgets allow changes of one to eight gray levels per pixel.

\begin{figure}
    \centering
    \includegraphics{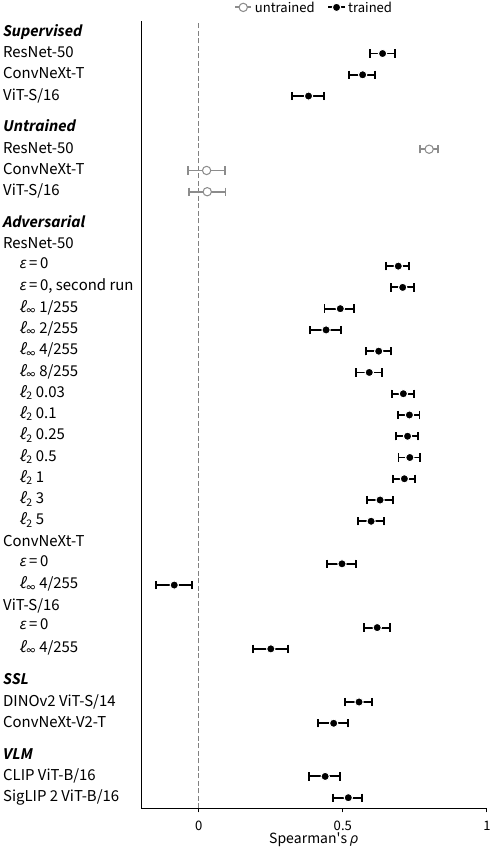}
    \caption{
        Spearman's rank correlation between the noise estimate $\hat\sigma$ and the departure scale $\eta^*_{10\%}$ under the uniform law for the 1000 ImageNet validation images.
        Encoders are listed as in Table~\ref{tab:models}.
        Bars show $95\%$ bootstrap intervals over images.
    }
    \label{fig:noise-correlation}
\end{figure}

\subsection{Kernel directions}
\label{app:kernel}

Section~\ref{sec:laws} introduced two direction laws, the uniform law and the singular laws $\mu_{(k)}$, each of them uniform on the unit sphere of a linear subspace of the perturbation space.
Here we describe a third such law, supported on the kernel of $\Jac$.

The singular vectors $\vv_k$ with $\lambda_k > 0$ span the row space of $\Jac$, whose orthogonal complement is the kernel $\ker \Jac$.
Since there are only $\dz$ vectors $\vv_k$, the row space has dimension at most $\dz$, leaving at least $\di - \dz$ dimensions for the kernel.
For example, a ResNet-50 produces 2048-dimensional embeddings; at ImageNet resolution, $\di = 3 \cdot 224^2$, so the kernel accounts for at least 98.6\% of the input dimensions, and the fraction is larger for encoders with lower embedding dimensionality.
Along any direction $\uu \in \ker \Jac$, the linear prediction is $\eta \Jac \uu = \boldsymbol{0}$ at every scale. For most dimensions of the input space, the Jacobian predicts that the embedding does not move at all.
Nothing distinguishes one kernel direction from another, and we again take $\mu$ uniform, now on the unit sphere of $\ker \Jac$.
Under this law, $\Jac \uu = \boldsymbol{0}$ for every $\uu$, so neither the size ratio nor the mean cosine is defined, and any nonzero response is superlinear.
We therefore compare the mean squared length of the response per unit scale with the level predicted for a typical direction, that is, with the value of $m_0(\xx)$ under the uniform law; formally, we define the \textit{kernel ratio}:
\[
    r^{\ker}_\eta(\xx) = \frac
    { \E_{\uu \sim \mu} \norm{ \Delta_\eta(\xx, \uu) }^2 / \eta^2 }
    { \norm{ \Jac }_F^2 / \di } \, .
\]
The kernel ratio vanishes as $\eta \to 0$ by construction.
The \textit{kernel scale} $\eta^{\ker}(\xx) = \inf \{ \eta > 0 \mid r^{\ker}_\eta(\xx) \ge 1 \}$ is then the scale at which the response along the kernel reaches the level predicted for a typical direction. 
When no such scale exists, we set $\eta^{\ker}(\xx) = + \infty$.

For each image, we take the first $64$ Gaussian vectors drawn for the uniform law (Appendix~\ref{app:sampling}), project out the components along the singular vectors $\vv_k$ with $\lambda_k \ge 10^{-6} \lambda_1$, and normalize the remaining vectors to unit length.
The resulting directions are uniform on the unit sphere of the orthogonal complement of these $\vv_k$, and we measure the response along each direction and its antipode.
At the smallest scales, the response along these directions is comparable in magnitude to rounding error.
We retain a direction at a given scale only if its squared response exceeds three times the noise reference of Appendix~\ref{app:numerical-floor}, and retain the scale for an image only if at least half of its directions pass this test.
The smallest scale never passes the latter test, so the kernel ratio is reported from the second scale, $0.012$ gray levels.

Figure~\ref{fig:kernel-scale} shows the kernel scale of each encoder, as distributions over the 1000 ImageNet validation images.
Kernel directions are predicted not to move the embedding at all, yet in most images of trained encoders the response along them reaches the level predicted for a random direction, at around one gray level.

\subsection{DeepFool and ODIN}
\label{app:deepfool-odin}

\paragraph{ODIN (temperature).}
Section~\ref{sec:odin-deepfool} states that at $T = 1000$ the departure of $\log S$ from its first-order prediction originates in the encoder.
Let $z_1, \dots, z_K$ be the logits, with $z_{\max}$ their maximum and $\bar z$ their mean.
Expanding the softmax in $1/T$ gives
\[
    \log S = \frac{z_{\max} - \bar z}{T} - \log K - \frac{1}{2 T^2} \cdot \frac{1}{K} \sum_{j=1}^{K} (z_j - \bar z)^2 + O(T^{-3}) \, .
\]
The first term is linear in the logits while the predicted class stays the same, and the logits are an affine function of the embedding.
At $T = 1000$, the remaining terms have a negligible effect on the change in $\log S$ that we measure.
At ODIN's step, the departure of $(z_{\max} - \bar z)/T$ from its first-order prediction matches that of $\log S$. The median ratio of the two departures is $1$ for each of the four models on both image sets.
The change of $\log S$ along the step is then the response $\Delta_\eta$ of the encoder projected on a fixed vector, and its first-order prediction is the same projection of $\eta \Jac \uu$.

\begin{figure*}[t]
    \centering
    \includegraphics{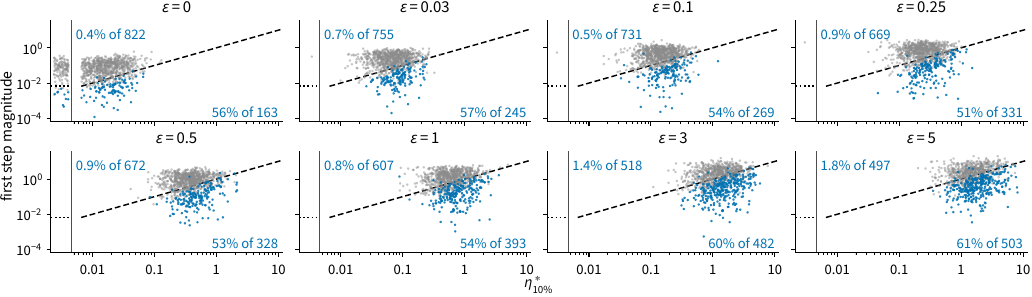}
    \caption{
        Length of DeepFool's first step against the departure scale $\eta^*_{10\%}$ along its direction, for ResNet-50 trained adversarially with $\ell_2$ budgets $\varepsilon$; $\varepsilon = 0$ denotes the model without adversarial training.
        Points, lines and percentages are as in Fig.~\ref{fig:deepfool}, right.
    }
    \label{fig:deepfool-l2}
\end{figure*}

\paragraph{DeepFool (adversarially trained models).}
We extend the measurements of Section~\ref{sec:odin-deepfool} to a family of adversarially trained ResNet-50 encoders with $\ell_2$ budgets~\cite{salman2020adversarially}.
Figure~\ref{fig:deepfool-l2} shows that, at every budget, the diagonal nearly separates the two outcomes of DeepFool's first step.
Above it, where the step is longer than the departure scale, the label almost never changes.
Below it, the images where the label does not change lie mostly close to the diagonal, and for steps well below the departure scale, the label changes in almost every image.
We conclude that whenever DeepFool's first step lies robustly within the linear regime, it succeeds in switching the label, while for larger steps more iterations are needed.

\begin{figure}
    \centering
    \includegraphics[width=\linewidth]{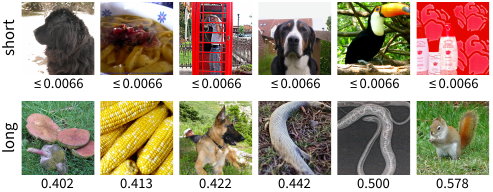}
    \caption{
        Images with the shortest and longest departure scales $\eta^*_{10\%}$ for the trained ResNet-50.
        The rows show six images drawn at random from the bottom 10\% (labeled \emph{short}) and top 10\% (\emph{long}) of the 1000 ImageNet validation images ranked by departure scale, with each image's $\eta^*_{10\%}$ shown below it.
        For images marked $\leq 0.0066$, $r_\eta$ or $c_\eta$ already deviates from its linear prediction by more than 10\% at the smallest measured scale.
        The short-scale images contain large low-frequency regions, the long-scale images fine texture (grass, corn kernels, foliage), consistent with the correlation between $\eta^*$ and estimated image noise in Fig.~\ref{fig:noise-correlation}.
    }
    \label{fig:radius-examples}
\end{figure}

\subsection{Speed and steadiness along the ray}
\label{app:speed-steadiness}
We compute the velocity $\ww_{\uu}(t) = \Jac(\xx + t\uu)\,\uu$ in double precision using a forward-mode Jacobian-vector product at $t = 0$ and at the 17 perturbation scales.
The images are every fourth of the 1000 ImageNet validation images in filename order, and the directions are the first 256 drawn for the uniform law, with their antipodes.
We evaluate the integral in $m_\eta$ using the trapezoidal rule, with nodes at $t = 0$ and at the measured scales up to $\eta$.

The split $r_\eta = \sigma_\eta \tau_\eta$ is exact only if its three terms are computed from the same directions and summarized in the same way.
For this reason, the size ratio in Fig.~\ref{fig:velocity} is recomputed from these 512 directions, with the mean of $\norm{\Jac\uu}^2$ over them in place of $m_0(\xx)$, and steadiness is the size ratio divided by the speed factor.
For the same 250 images, this size ratio agrees with the estimate from all 2048 directions and the exact $m_0(\xx)$ to within $0.33\%$ at every scale for all four encoders.

The curves in Fig.~\ref{fig:velocity} are geometric means over the 250 images, so the product of the curves for the speed factor and steadiness equals the size ratio curve.
Figure~\ref{fig:measurement-profiles} shows medians over the 1000 images, so its curves are not identical to the size ratios of Fig.~\ref{fig:velocity}.
Across the 17 scales, the maximum difference between the geometric mean and the median ranges from $16\%$ to $38\%$ across encoders, while the maximum difference between estimates from 250 and 1000 images ranges from $7\%$ to $11\%$.
For ResNet-50, the peak size ratio is $1.30$ in both figures, at adjacent measured scales. The two peak values are $4.77$ and $4.75$ for ConvNeXt-T, and $2.51$ and $2.92$ for ViT-S/16.
For DINOv2, Fig.~\ref{fig:measurement-profiles} shows the class token and Fig.~\ref{fig:velocity} the head input (Table~\ref{tab:models}), so the two curves refer to different embeddings.

The normalized inner product used in Section~\ref{sec:velocity-exp} is
\[
    \frac{\E_{\uu \sim \mu} \inner{\ww_\uu(\eta)}{\ww_\uu(0)}}{\sqrt{\E_{\uu \sim \mu} \norm{\ww_\uu(\eta)}^2 \; \E_{\uu \sim \mu} \norm{\ww_\uu(0)}^2}} \, ,
\]
which we compute for each image from the 512 directions and summarize by the median over the 250 images.
It equals $1$ only if the velocity at $\eta$ is the same positive multiple of the starting velocity in every direction.
A value below $1$ can come from the velocity turning, or from its speed changing by a different factor in different directions.
To separate the two, we also compute the cosine between $\ww_\uu(\eta)$ and $\ww_\uu(0)$ in each direction and average it over the 512 directions.
Its median over the images differs from that of the normalized inner product by at most $0.014$ at every scale for ResNet-50, ConvNeXt-T and DINOv2, and by at most $0.10$ for ViT-S/16.
It is below $0.5$ from $0.31$ to $0.36$ gray levels on in the three supervised encoders and from $0.61$ on in DINOv2, and below $0.1$ beyond $1.0$ to $2.7$ gray levels.

\subsection{Other image sets}
\label{app:other-image-sets}
We ask whether the linear regime is also short on ImageNet-O, ImageNet-A and generated images.
We repeat the uniform-law measurement of Fig.~\ref{fig:departure_distributions} on three other image sets for the three supervised encoders (Table~\ref{tab:datasets}).

Each set contains $256$ images.
For the first two sets, we sample images at random from ImageNet-O and ImageNet-A~\cite{hendrycks2021natural}, respectively.
The third set consists of images generated with Stable Diffusion~1.4~\cite{rombach2022high}, one for each of $256$ ImageNet classes, using the class name and its gloss as the prompt.
As a reference, we also evaluate $256$ ImageNet validation images, one from each of the same classes.
All images are resized and center-cropped to $224 \times 224$.

The median departure scale stays below one gray level across all image sets at both tolerances.
At a tolerance of $10\%$, the median departure scale lies between $0.07$ and $0.20$ gray levels across all sets and encoders.
For the two convolutional encoders, the shortest median scales occur on ImageNet-O. They are $0.09$ for ResNet-50 and $0.07$ for ConvNeXt-T, compared with $0.14$ and $0.12$ on the validation images.
For ViT-S/16, the longest median occurs on the Stable Diffusion images, at $0.20$ compared with $0.14$ on the validation set.
At a tolerance of $50\%$, the median remains below one gray level on every image set.

\begin{table*}[t]
    \centering
    \caption{
        Departure scales under the uniform law, in gray levels, for $256$ images from each of the ImageNet validation set (IM-VAL), ImageNet-O (IM-O), ImageNet-A (IM-A) and the Stable Diffusion set (SD).
        Columns $r_\eta$ and $c_\eta$ report the departure scales for the size ratio and mean cosine separately. $\eta^*$ is the smaller of the two.
        Cells report medians over images, with interquartile ranges.
        Values in italics lie at or below the smallest measured scale.
    }
    \input{tables/datasets}
    \label{tab:datasets}
\end{table*}

\subsection{Shortest and longest departure scales}
\label{app:shortest-longest}
We ask whether images with finer texture tend to have longer departure scales.
We relate the departure scale of an image to an estimate of its noise level.

Figure~\ref{fig:radius-examples} shows examples from the shortest and longest departure-scale deciles for the trained ResNet-50.
We use a wavelet-based noise estimate~\cite{donoho1994ideal} to compare the amount of fine texture across images.
The estimate is $\hat\sigma = \operatorname{median}(|d|) / 0.6745$, where $d$ contains the diagonal detail coefficients from a one-level wavelet transform of the image luminance using the Daubechies wavelet with four vanishing moments.
The estimate is designed for images with added noise.
On our images it grows with the amount of fine texture.

Figure~\ref{fig:noise-correlation} shows Spearman's rank correlation $\rho$ between $\hat\sigma$ and $\eta^*_{10\%}$ under the uniform law for the 1000 ImageNet validation images.
For the supervised, self-supervised and vision--language encoders, $\rho$ lies between $0.38$ and $0.64$. Images with larger noise estimates tend to have longer departure scales.
For adversarially trained ResNet-50, the correlation ranges from $0.44$ to $0.73$ across the tested budgets.
With an $\ell_\infty$ budget of $4/255$, the correlation falls from $0.62$ to $0.25$ for ViT-S/16 and from $0.50$ to $-0.09$ for ConvNeXt-T.
Among the untrained encoders, ResNet-50 has $\rho = 0.80$, the highest correlation observed, and ConvNeXt-T and ViT-S/16 have $\rho = 0.03$.
The correlation of ResNet-50 is already present at initialization, and that of the other two appears with training.

\subsection{Other layers}
\label{app:other-layers}
We ask whether the size ratio and the mean cosine depend on depth in the same way.
We repeat the uniform-law measurement at different layers of five encoders, treating the output of each layer as the embedding (Fig.~\ref{fig:depth_rc}).
For the adversarially trained ResNet-50, we use the checkpoint from the robustness library~\cite{engstrom2019robustness} with an $\ell_2$ budget of $\varepsilon = 3$, rather than the checkpoint in Table~\ref{tab:models}.

At the input, both statistics equal $1$ at every scale.
In the four encoders without adversarial training, the mean cosine falls at similar perturbation scales across all layers beyond the first blocks.
The size ratio behaves differently.
At intermediate layers it falls below $1$ at large scales, and a superlinear response with $r_\eta > 2$ appears only in the last layers of ConvNeXt-T, ViT-S/16 and DINOv2.
In the adversarially trained ResNet-50, the size ratio rises above $1$ at every depth beyond the stem and grows with depth.

\subsection{Reference scales}
\label{app:reference-scales}
The reference scales in Fig.~\ref{fig:departure_distributions} are expressed as root-mean-square changes per pixel value in gray levels.
This unit is independent of image resolution.

The 8-bit rounding error is $1/\sqrt{12} \approx 0.29$ gray levels, the root-mean-square error of rounding to steps of one gray level~\cite{bennett1948spectra}.
ODIN~\cite{liang2018enhancing} takes a step of $\varepsilon = 0.0014$ per pixel value on the $[0, 1]$ scale, which is $255 \cdot 0.0014 = 0.357$ gray levels.
This is the step used in the original paper, the default in OpenOOD~v1.5~\cite{zhang2024openood}, and the value we use in Section~\ref{sec:odin-deepfool}.
FGSM~\cite{goodfellow2015explaining} is drawn at $\varepsilon = 1/255$, one gray level, which its authors describe as the smallest step of an 8-bit image encoding.
The $\ell_\infty$ attack threshold of~\cite{simongabriel2019first} is $0.5\%$ of the pixel range, which is $255 \cdot 0.005 = 1.275$ gray levels.
Integrated Gradients~\cite{sundararajan2017axiomatic} integrates along the straight path from a black image to $\xx$, dividing it into $m$ equal segments of displacement $\xx / m$.
With $m = 50$, as in the authors' code, a segment has a median scale of $2.60$ gray levels across our 1000 images.
Free adversarial training~\cite{shafahi2019adversarial} uses an $\ell_\infty$ budget of $4$ gray levels for ResNet-50 on ImageNet.

ODIN and FGSM step along the sign of a gradient, so every pixel value changes by $\varepsilon$, and the root-mean-square change equals $\varepsilon$.
For the two $\ell_\infty$ budgets~\cite{simongabriel2019first, shafahi2019adversarial}, the root-mean-square change is bounded above by the budget.
At ImageNet resolution, the largest absolute pixel change under a uniformly random perturbation is about $\sqrt{2 \ln \di} \approx 5$ times its scale in gray levels, so a departure scale of $0.15$ gray levels corresponds to an $\ell_\infty$ change of about $0.7$ gray levels.

The departure scales of Fig.~\ref{fig:departure_distributions} are measured along uniformly random directions and along the leading singular direction, and not along the directions used in these papers.
The uniform law is a reference that depends on no method.
Along singular directions the linear regime is shorter than under it.
For each of the seven trained encoders in Fig.~\ref{fig:departure_distributions}, the median $\eta^*_{10\%}$ at singular ranks $1$, $16$ and $256$ (Table~\ref{tab:radii_rank}) lies below the uniform-law median (Table~\ref{tab:radii_uniform}). At rank $1$, the uniform-law median is at least four times the singular-law median.
ODIN, FGSM and the adversarial perturbations of~\cite{simongabriel2019first, shafahi2019adversarial} follow a gradient or its sign.
A gradient of any function of the embedding has the form $\Jac^\top \mathbf{a} = \sum_k \sqrt{\lambda_k} \, \inner{\mathbf{e}_k}{\mathbf{a}} \, \vv_k$, a combination of the right singular vectors.
Along the directions of ODIN and DeepFool themselves, the median departure scale is $0.015$ to $0.13$ and at most $0.04$ gray levels, respectively (Section~\ref{sec:odin-deepfool}).

%% file: tables/datasets.tex
\begingroup
\small
\setlength{\tabcolsep}{2.5pt}
\renewcommand{\arraystretch}{1.32}
\begin{tabular}{@{}c@{\hspace{6pt}}l@{\hspace{8pt}}r@{\,}lr@{\,}lr@{\,}l@{\hspace{12pt}}r@{\,}lr@{\,}lr@{\,}l@{}}
  \toprule
    & & \multicolumn{6}{c}{$\epsilon = 10\%$} & \multicolumn{6}{c}{$\epsilon = 50\%$} \\
  \cmidrule(lr){3-8}\cmidrule(l){9-14}
  & \textbf{Encoder} & \multicolumn{2}{c}{$\eta^\ast$} & \multicolumn{2}{c}{$r_\eta$} & \multicolumn{2}{c}{$c_\eta$} & \multicolumn{2}{c}{$\eta^\ast$} & \multicolumn{2}{c}{$r_\eta$} & \multicolumn{2}{c}{$c_\eta$} \\
  \midrule
  \multirow{3}{*}{\rotatebox[origin=c]{90}{\textbf{\textit{IM-VAL}}}} & ResNet-50 & 0.14 & [0.07, 0.23] & 0.36 & [0.13, 0.77] & 0.14 & [0.07, 0.23] & 0.79 & [0.49, 1.20] & 6.71 & [1.13, 15.05] & 0.79 & [0.49, 1.20] \\
   & ConvNeXt-T & 0.12 & [0.07, 0.19] & 0.12 & [0.07, 0.19] & 0.16 & [0.09, 0.25] & 0.41 & [0.21, 0.70] & 0.41 & [0.21, 0.70] & 0.63 & [0.36, 1.01] \\
   & ViT-S/16 & 0.14 & [0.09, 0.24] & 0.16 & [0.10, 0.36] & 0.19 & [0.13, 0.30] & 0.68 & [0.41, 1.04] & 0.83 & [0.46, 2.20] & 0.80 & [0.54, 1.13] \\
  \addlinespace[2pt]\arrayrulecolor{black!60}\cmidrule{1-14}\arrayrulecolor{black}\addlinespace[2pt]
  \multirow{3}{*}{\rotatebox[origin=c]{90}{\textbf{\textit{IM-O}}}} & ResNet-50 & 0.09 & [\textit{0.01}, 0.18] & 0.19 & [\textit{0.01}, 0.50] & 0.09 & [\textit{0.01}, 0.18] & 0.47 & [0.23, 0.90] & 1.14 & [0.31, 8.86] & 0.48 & [0.23, 0.90] \\
   & ConvNeXt-T & 0.07 & [0.03, 0.12] & 0.07 & [0.03, 0.12] & 0.09 & [0.04, 0.16] & 0.21 & [0.10, 0.40] & 0.21 & [0.10, 0.40] & 0.35 & [0.17, 0.63] \\
   & ViT-S/16 & 0.14 & [0.08, 0.25] & 0.16 & [0.08, 0.29] & 0.16 & [0.12, 0.28] & 0.55 & [0.35, 0.93] & 0.64 & [0.36, 1.48] & 0.67 & [0.43, 1.09] \\
  \addlinespace[2pt]\arrayrulecolor{black!60}\cmidrule{1-14}\arrayrulecolor{black}\addlinespace[2pt]
  \multirow{3}{*}{\rotatebox[origin=c]{90}{\textbf{\textit{IM-A}}}} & ResNet-50 & 0.12 & [0.07, 0.20] & 0.28 & [0.14, 0.52] & 0.12 & [0.07, 0.20] & 0.60 & [0.41, 0.90] & 1.93 & [0.80, 8.01] & 0.61 & [0.42, 0.90] \\
   & ConvNeXt-T & 0.08 & [0.05, 0.13] & 0.08 & [0.05, 0.13] & 0.12 & [0.08, 0.18] & 0.26 & [0.17, 0.44] & 0.26 & [0.17, 0.44] & 0.44 & [0.28, 0.71] \\
   & ViT-S/16 & 0.16 & [0.09, 0.27] & 0.18 & [0.09, 0.34] & 0.20 & [0.13, 0.34] & 0.71 & [0.37, 1.12] & 0.74 & [0.40, 1.42] & 0.82 & [0.52, 1.34] \\
  \addlinespace[2pt]\arrayrulecolor{black!60}\cmidrule{1-14}\arrayrulecolor{black}\addlinespace[2pt]
  \multirow{3}{*}{\rotatebox[origin=c]{90}{\textbf{\textit{SD}}}} & ResNet-50 & 0.17 & [0.13, 0.23] & 0.39 & [0.24, 0.60] & 0.17 & [0.13, 0.23] & 0.78 & [0.61, 1.07] & 1.62 & [0.85, 9.31] & 0.79 & [0.61, 1.07] \\
   & ConvNeXt-T & 0.10 & [0.07, 0.14] & 0.10 & [0.07, 0.14] & 0.13 & [0.09, 0.20] & 0.31 & [0.21, 0.45] & 0.31 & [0.21, 0.45] & 0.51 & [0.34, 0.80] \\
   & ViT-S/16 & 0.20 & [0.12, 0.29] & 0.22 & [0.12, 0.35] & 0.25 & [0.16, 0.38] & 0.79 & [0.54, 1.18] & 0.91 & [0.58, 1.65] & 0.95 & [0.68, 1.47] \\
  \bottomrule
\end{tabular}
\endgroup

%% file: sections/appendix/C.tex
\section{Tables}

The following tables report departure scales at tolerances of $10\%$ and $50\%$, including the numerical values underlying Fig.~\ref{fig:departure_distributions}.
Table~\ref{tab:radii_uniform} gives the scales under the uniform law for the encoders in the figure, as well as the untrained and adversarially trained encoders.
Table~\ref{tab:radii_rank} covers singular directions of ranks $1$, $16$ and $256$, and Table~\ref{tab:radii_mem} covers the fine-tuned encoders of Appendix~\ref{app:effect-of-training}.

\begin{table*}
    \centering
    \caption{
        Departure scales under the uniform law, in gray levels, for the 1000 ImageNet validation images.
        Columns $r_\eta$ and $c_\eta$ report the departure scales for the size ratio and mean cosine separately. $\eta^*$ is the smaller of the two.
        Cells report medians over images, with interquartile ranges.
        Italics indicate values at or below the smallest measured scale. A dash indicates no departure within the measured range.
        Encoders are listed as in Table~\ref{tab:models}.
    }
    \input{tables/radii_uniform}
    \label{tab:radii_uniform}
\end{table*}

\begin{table*}
    \centering
    \caption{
        Departure scales along the singular directions of rank $k$, in gray levels, for the 1000 ImageNet validation images.
        The column definitions and reporting conventions (including italics and dashes) follow Table~\ref{tab:radii_uniform}.
    }
    \input{tables/radii_rank}
    \label{tab:radii_rank}
\end{table*}

\begin{table*}
    \centering
    \caption{
        Departure scales of the three supervised encoders under the uniform law and along the singular directions of rank $k$, at their pretrained checkpoints and after fine-tuning on the 1000 measured ImageNet validation images with true or shuffled labels.
        The column definitions and reporting conventions (including italics and dashes) follow Table~\ref{tab:radii_uniform}.
    }
    \resizebox{\textwidth}{!}{\input{tables/radii_mem}}
    \label{tab:radii_mem}
\end{table*}

%% file: tables/radii_uniform.tex
\begingroup
\footnotesize
\setlength{\tabcolsep}{2.5pt}
\renewcommand{\arraystretch}{1.32}
\begin{tabular}{@{}c@{\hspace{6pt}}l@{\hspace{8pt}}r@{\,}lr@{\,}lr@{\,}l@{\hspace{12pt}}r@{\,}lr@{\,}lr@{\,}l@{}}
  \toprule
    & & \multicolumn{6}{c}{$\epsilon = 10\%$} & \multicolumn{6}{c}{$\epsilon = 50\%$} \\
  \cmidrule(lr){3-8}\cmidrule(l){9-14}
  & \textbf{Encoder} & \multicolumn{2}{c}{$\eta^\ast$} & \multicolumn{2}{c}{$r_\eta$} & \multicolumn{2}{c}{$c_\eta$} & \multicolumn{2}{c}{$\eta^\ast$} & \multicolumn{2}{c}{$r_\eta$} & \multicolumn{2}{c}{$c_\eta$} \\
  \midrule
  \multirow{3}{*}{\rotatebox[origin=c]{90}{\textbf{\textit{Supervised}}}} & ResNet-50 & 0.15 & [0.08, 0.23] & 0.39 & [0.15, 0.81] & 0.15 & [0.08, 0.24] & 0.82 & [0.52, 1.22] & 6.50 & [1.18, 14.42] & 0.82 & [0.52, 1.22] \\
   & ConvNeXt-T & 0.12 & [0.07, 0.21] & 0.12 & [0.07, 0.21] & 0.16 & [0.10, 0.26] & 0.40 & [0.22, 0.71] & 0.40 & [0.22, 0.72] & 0.62 & [0.36, 1.03] \\
   & ViT-S/16 & 0.16 & [0.09, 0.26] & 0.18 & [0.10, 0.37] & 0.21 & [0.14, 0.33] & 0.70 & [0.43, 1.13] & 0.95 & [0.48, 2.20] & 0.84 & [0.55, 1.30] \\
  \addlinespace[2pt]\arrayrulecolor{black!60}\cmidrule{1-14}\arrayrulecolor{black}\addlinespace[2pt]
  \multirow{3}{*}{\rotatebox[origin=c]{90}{\textbf{\textit{Untrained}}}} & ResNet-50 & 0.26 & [0.16, 0.39] & 0.29 & [0.16, 0.45] & 0.26 & [0.16, 0.39] & 0.83 & [0.49, 1.29] & 0.83 & [0.49, 1.29] & 1.03 & [0.65, 1.50] \\
   & ConvNeXt-T & 13.16 & [7.75, 23.87] & 13.78 & [7.85, 26.60] & 29.52 & [16.29, 56.88] & --- & [42.40, ---] & --- & [42.50, ---] & --- & [---, ---] \\
   & ViT-S/16 & 6.94 & [4.87, 9.21] & 6.96 & [4.87, 9.52] & 8.98 & [6.83, 12.15] & 25.69 & [17.81, 47.13] & 26.32 & [17.82, ---] & 43.32 & [30.95, ---] \\
  \addlinespace[2pt]\arrayrulecolor{black!60}\cmidrule{1-14}\arrayrulecolor{black}\addlinespace[2pt]
  \multirow{20}{*}{\rotatebox[origin=c]{90}{\textbf{\textit{Adversarial}}}} & ResNet-50 & \multicolumn{12}{@{}l@{}}{{\color{black!25}\xleaders\hrule height 0.62ex depth -0.55ex\hskip 0pt plus 1filll}} \\
   & \quad $\varepsilon = 0$ & 0.23 & [0.13, 0.37] & 0.38 & [0.20, 0.73] & 0.23 & [0.14, 0.37] & 1.03 & [0.61, 1.62] & 1.37 & [0.68, 5.20] & 1.05 & [0.65, 1.62] \\
   & \quad $\varepsilon = 0$, second run & 0.23 & [0.14, 0.36] & 0.40 & [0.22, 0.74] & 0.23 & [0.14, 0.36] & 1.03 & [0.67, 1.61] & 1.45 & [0.77, 10.63] & 1.04 & [0.69, 1.61] \\
   & \quad $\ell_\infty$ 1/255 & 0.46 & [0.38, 0.64] & 0.46 & [0.38, 0.64] & 0.64 & [0.49, 0.79] & 1.43 & [1.18, 1.89] & 1.43 & [1.18, 1.89] & 2.34 & [1.89, 2.97] \\
   & \quad $\ell_\infty$ 2/255 & 0.43 & [0.36, 0.56] & 0.43 & [0.36, 0.56] & 0.60 & [0.46, 0.75] & 1.39 & [1.13, 1.74] & 1.39 & [1.13, 1.74] & 2.24 & [1.80, 2.78] \\
   & \quad $\ell_\infty$ 4/255 & 0.38 & [0.21, 0.51] & 0.38 & [0.21, 0.51] & 0.50 & [0.35, 0.70] & 1.29 & [0.96, 1.68] & 1.29 & [0.96, 1.68] & 2.08 & [1.61, 2.71] \\
   & \quad $\ell_\infty$ 8/255 & 0.09 & [\textit{0.01}, 0.21] & 0.11 & [0.02, 0.23] & 0.12 & [\textit{0.01}, 0.26] & 0.59 & [0.35, 0.85] & 0.70 & [0.45, 1.01] & 0.83 & [0.47, 1.26] \\
   & \quad $\ell_2$ 0.03 & 0.23 & [0.11, 0.42] & 0.23 & [0.11, 0.42] & 0.27 & [0.14, 0.47] & 0.72 & [0.38, 1.33] & 0.72 & [0.38, 1.33] & 1.13 & [0.65, 1.92] \\
   & \quad $\ell_2$ 0.1 & 0.20 & [0.09, 0.35] & 0.20 & [0.09, 0.35] & 0.26 & [0.13, 0.44] & 0.63 & [0.33, 1.08] & 0.63 & [0.33, 1.08] & 1.06 & [0.60, 1.76] \\
   & \quad $\ell_2$ 0.25 & 0.21 & [0.12, 0.35] & 0.21 & [0.12, 0.35] & 0.28 & [0.16, 0.45] & 0.68 & [0.39, 1.06] & 0.68 & [0.39, 1.06] & 1.14 & [0.69, 1.77] \\
   & \quad $\ell_2$ 0.5 & 0.21 & [0.12, 0.32] & 0.21 & [0.12, 0.32] & 0.29 & [0.18, 0.43] & 0.68 & [0.43, 0.97] & 0.68 & [0.43, 0.97] & 1.16 & [0.77, 1.70] \\
   & \quad $\ell_2$ 1 & 0.22 & [0.13, 0.29] & 0.22 & [0.13, 0.29] & 0.29 & [0.20, 0.42] & 0.68 & [0.44, 0.92] & 0.68 & [0.44, 0.92] & 1.17 & [0.78, 1.62] \\
   & \quad $\ell_2$ 3 & 0.23 & [0.16, 0.31] & 0.23 & [0.16, 0.31] & 0.33 & [0.23, 0.43] & 0.72 & [0.52, 0.95] & 0.72 & [0.52, 0.95] & 1.26 & [0.92, 1.66] \\
   & \quad $\ell_2$ 5 & 0.22 & [0.15, 0.29] & 0.22 & [0.15, 0.29] & 0.31 & [0.23, 0.41] & 0.69 & [0.51, 0.89] & 0.69 & [0.51, 0.89] & 1.19 & [0.90, 1.58] \\
   & ConvNeXt-T & \multicolumn{12}{@{}l@{}}{{\color{black!25}\xleaders\hrule height 0.62ex depth -0.55ex\hskip 0pt plus 1filll}} \\
   & \quad $\varepsilon = 0$ & 0.13 & [0.07, 0.21] & 0.13 & [0.07, 0.22] & 0.16 & [0.09, 0.26] & 0.43 & [0.21, 0.74] & 0.44 & [0.21, 0.83] & 0.62 & [0.33, 1.00] \\
   & \quad $\ell_\infty$ 4/255 & 0.42 & [0.35, 0.49] & 0.42 & [0.35, 0.49] & 0.59 & [0.48, 0.71] & 1.29 & [1.06, 1.52] & 1.29 & [1.06, 1.52] & 2.17 & [1.80, 2.62] \\
   & ViT-S/16 & \multicolumn{12}{@{}l@{}}{{\color{black!25}\xleaders\hrule height 0.62ex depth -0.55ex\hskip 0pt plus 1filll}} \\
   & \quad $\varepsilon = 0$ & 0.23 & [0.13, 0.40] & 0.32 & [0.14, 0.78] & 0.26 & [0.16, 0.44] & 1.06 & [0.56, 1.80] & 5.22 & [1.14, 12.04] & 1.11 & [0.61, 1.83] \\
   & \quad $\ell_\infty$ 4/255 & 0.77 & [0.66, 0.97] & 0.77 & [0.66, 0.97] & 1.13 & [0.89, 1.36] & 2.39 & [2.01, 2.98] & 2.39 & [2.01, 2.98] & 4.04 & [3.30, 5.09] \\
  \addlinespace[2pt]\arrayrulecolor{black!60}\cmidrule{1-14}\arrayrulecolor{black}\addlinespace[2pt]
  \multirow{2}{*}{\rotatebox[origin=c]{90}{\textbf{\textit{SSL}}}} & DINOv2 ViT-S/14 & 0.31 & [0.21, 0.49] & 0.31 & [0.21, 0.50] & 0.41 & [0.27, 0.66] & 1.23 & [0.71, 2.16] & 1.25 & [0.71, 3.08] & 1.67 & [1.10, 2.66] \\
   & ConvNeXt-V2-T & 0.15 & [0.08, 0.29] & 0.15 & [0.08, 0.30] & 0.19 & [0.10, 0.32] & 0.74 & [0.27, 1.41] & 1.01 & [0.27, 48.49] & 0.85 & [0.41, 1.43] \\
  \addlinespace[2pt]\arrayrulecolor{black!60}\cmidrule{1-14}\arrayrulecolor{black}\addlinespace[2pt]
  \multirow{2}{*}{\rotatebox[origin=c]{90}{\textbf{\textit{VLM}}}} & CLIP ViT-B/16 & 0.05 & [0.04, 0.09] & 0.05 & [0.04, 0.09] & 0.07 & [0.05, 0.12] & 0.17 & [0.12, 0.33] & 0.17 & [0.12, 0.35] & 0.25 & [0.17, 0.41] \\
   & SigLIP 2 ViT-B/16 & 0.17 & [0.12, 0.26] & 0.21 & [0.12, 0.40] & 0.19 & [0.13, 0.27] & 0.60 & [0.38, 0.90] & 0.81 & [0.42, 3.86] & 0.63 & [0.43, 0.92] \\
  \bottomrule
\end{tabular}
\endgroup

%% file: tables/radii_rank.tex
\begingroup
\scriptsize
\setlength{\tabcolsep}{2.5pt}
\begin{tabular}{@{}c@{\hspace{6pt}}l@{\hspace{6pt}}r@{\hspace{8pt}}r@{\,}lr@{\,}lr@{\,}l@{\hspace{12pt}}r@{\,}lr@{\,}lr@{\,}l@{}}
  \toprule
    & & & \multicolumn{6}{c}{$\epsilon = 10\%$} & \multicolumn{6}{c}{$\epsilon = 50\%$} \\
  \cmidrule(lr){4-9}\cmidrule(l){10-15}
  & \textbf{Encoder} & $k$ & \multicolumn{2}{c}{$\eta^\ast$} & \multicolumn{2}{c}{$r_\eta$} & \multicolumn{2}{c}{$c_\eta$} & \multicolumn{2}{c}{$\eta^\ast$} & \multicolumn{2}{c}{$r_\eta$} & \multicolumn{2}{c}{$c_\eta$} \\
  \midrule
  \multirow{9}{*}{\rotatebox[origin=c]{90}{\textbf{\textit{Supervised}}}} & ResNet-50 & 1 & \textit{0.007} & [\textit{0.007}, 0.008] & \textit{0.007} & [\textit{0.007}, 0.008] & 0.064 & [0.014, 0.146] & 0.034 & [0.012, 0.060] & 0.034 & [0.012, 0.060] & 2.490 & [0.589, 5.046] \\
   &  & 16 & 0.007 & [\textit{0.007}, 0.012] & 0.007 & [\textit{0.007}, 0.012] & 0.059 & [0.029, 0.100] & 0.067 & [0.040, 0.100] & 0.068 & [0.040, 0.100] & 1.266 & [0.646, 2.099] \\
   &  & 256 & 0.019 & [0.008, 0.030] & 0.046 & [0.015, 0.128] & 0.020 & [0.009, 0.031] & 0.195 & [0.095, 0.302] & 0.762 & [0.406, 1.369] & 0.195 & [0.096, 0.303] \\
  \addlinespace[1.5pt]
   & ConvNeXt-T & 1 & 0.027 & [0.014, 0.047] & 0.027 & [0.014, 0.047] & 0.107 & [0.052, 0.189] & 0.099 & [0.053, 0.172] & 0.099 & [0.053, 0.172] & 5.142 & [2.201, 10.037] \\
   &  & 16 & 0.075 & [0.052, 0.107] & 0.078 & [0.056, 0.123] & 0.134 & [0.090, 0.210] & 0.327 & [0.231, 0.488] & 0.329 & [0.231, 0.494] & 2.849 & [1.710, 4.368] \\
   &  & 256 & 0.075 & [0.046, 0.123] & 0.075 & [0.046, 0.124] & 0.095 & [0.062, 0.138] & 0.401 & [0.181, 0.757] & 0.519 & [0.181, 2.576] & 0.496 & [0.282, 0.789] \\
  \addlinespace[1.5pt]
   & ViT-S/16 & 1 & \textit{0.007} & [\textit{0.007}, \textit{0.007}] & \textit{0.007} & [\textit{0.007}, \textit{0.007}] & \textit{0.007} & [\textit{0.007}, 0.010] & \textit{0.007} & [\textit{0.007}, 0.013] & \textit{0.007} & [\textit{0.007}, 0.014] & 0.014 & [\textit{0.007}, 0.204] \\
   &  & 16 & 0.017 & [0.007, 0.049] & 0.019 & [0.007, 0.074] & 0.024 & [0.009, 0.068] & 0.372 & [0.189, 0.565] & 0.378 & [0.203, 0.572] & 1.271 & [0.733, 1.917] \\
   &  & 256 & 0.013 & [0.007, 0.024] & 0.013 & [0.007, 0.024] & 0.019 & [0.010, 0.037] & 0.039 & [0.021, 0.073] & 0.039 & [0.021, 0.073] & 0.071 & [0.036, 0.142] \\
  \addlinespace[2pt]\arrayrulecolor{black!60}\cmidrule{1-15}\arrayrulecolor{black}\addlinespace[2pt]
  \multirow{9}{*}{\rotatebox[origin=c]{90}{\textbf{\textit{Untrained}}}} & ResNet-50 & 1 & 0.138 & [0.104, 0.171] & 0.138 & [0.104, 0.171] & 11.911 & [7.102, 12.319] & 1.198 & [0.935, 1.447] & 1.198 & [0.935, 1.447] & 15.388 & [10.316, 15.863] \\
   &  & 16 & 0.072 & [0.057, 0.084] & 0.072 & [0.057, 0.084] & 1.905 & [1.456, 2.247] & 0.572 & [0.464, 0.678] & 0.572 & [0.464, 0.678] & 5.868 & [4.879, 6.923] \\
   &  & 256 & 0.078 & [0.063, 0.092] & 0.079 & [0.064, 0.092] & 0.461 & [0.383, 0.549] & 1.423 & [1.033, 1.884] & 6.530 & [1.107, 9.778] & 1.919 & [1.550, 2.290] \\
  \addlinespace[1.5pt]
   & ConvNeXt-T & 1 & 0.585 & [0.374, 0.997] & 0.585 & [0.374, 0.997] & 4.130 & [2.371, 7.069] & 2.132 & [1.312, 3.546] & 2.132 & [1.312, 3.546] & --- & [---, ---] \\
   &  & 16 & 0.845 & [0.516, 1.489] & 0.845 & [0.516, 1.489] & 5.890 & [3.400, 9.869] & 3.224 & [1.999, 5.409] & 3.224 & [1.999, 5.409] & --- & [---, ---] \\
   &  & 256 & 1.562 & [0.978, 2.648] & 1.562 & [0.978, 2.648] & 11.754 & [6.786, 19.732] & 6.162 & [3.849, 10.106] & 6.162 & [3.849, 10.106] & --- & [---, ---] \\
  \addlinespace[1.5pt]
   & ViT-S/16 & 1 & 1.252 & [0.742, 2.129] & 1.252 & [0.742, 2.129] & 3.758 & [2.264, 6.187] & 4.042 & [2.413, 6.809] & 4.042 & [2.413, 6.809] & --- & [---, ---] \\
   &  & 16 & 2.061 & [1.225, 3.057] & 2.061 & [1.225, 3.057] & 5.275 & [3.229, 8.927] & 6.524 & [3.959, 10.029] & 6.524 & [3.959, 10.029] & --- & [---, ---] \\
   &  & 256 & 4.584 & [2.551, 7.317] & 4.886 & [2.620, 8.816] & 4.921 & [2.872, 7.887] & 19.734 & [10.931, 32.588] & 28.800 & [11.350, ---] & 22.378 & [13.052, 35.093] \\
  \addlinespace[2pt]\arrayrulecolor{black!60}\cmidrule{1-15}\arrayrulecolor{black}\addlinespace[2pt]
  \multirow{6}{*}{\rotatebox[origin=c]{90}{\textbf{\textit{SSL}}}} & DINOv2 & 1 & 0.035 & [0.018, 0.062] & 0.036 & [0.019, 0.062] & 0.135 & [0.077, 0.231] & 0.123 & [0.069, 0.209] & 0.123 & [0.069, 0.209] & 3.883 & [1.793, 6.694] \\
   & ViT-S/14 & 16 & 0.118 & [0.076, 0.168] & 0.125 & [0.080, 0.184] & 0.220 & [0.126, 0.355] & 0.451 & [0.319, 0.646] & 0.455 & [0.320, 0.651] & 3.008 & [1.742, 4.604] \\
   &  & 256 & 0.074 & [0.044, 0.123] & 0.074 & [0.044, 0.123] & 0.102 & [0.066, 0.162] & 0.260 & [0.152, 0.473] & 0.260 & [0.152, 0.473] & 0.459 & [0.273, 0.747] \\
  \addlinespace[1.5pt]
   & ConvNeXt-V2-T & 1 & 0.013 & [\textit{0.007}, 0.037] & 0.013 & [\textit{0.007}, 0.038] & 0.037 & [0.014, 0.095] & 0.057 & [0.018, 0.123] & 0.062 & [0.020, 0.133] & 0.240 & [0.049, 4.507] \\
   &  & 16 & 0.025 & [0.012, 0.047] & 0.026 & [0.012, 0.061] & 0.030 & [0.016, 0.054] & 0.109 & [0.038, 0.241] & 0.252 & [0.038, 0.502] & 0.117 & [0.061, 0.264] \\
   &  & 256 & 0.016 & [0.008, 0.031] & 0.016 & [0.008, 0.031] & 0.023 & [0.013, 0.043] & 0.049 & [0.026, 0.095] & 0.049 & [0.026, 0.095] & 0.080 & [0.044, 0.153] \\
  \addlinespace[2pt]\arrayrulecolor{black!60}\cmidrule{1-15}\arrayrulecolor{black}\addlinespace[2pt]
  \multirow{6}{*}{\rotatebox[origin=c]{90}{\textbf{\textit{VLM}}}} & CLIP & 1 & \textit{0.007} & [\textit{0.007}, 0.007] & \textit{0.007} & [\textit{0.007}, 0.007] & 0.017 & [0.010, 0.026] & 0.017 & [0.010, 0.025] & 0.017 & [0.010, 0.025] & 0.179 & [0.078, 0.802] \\
   & ViT-B/16 & 16 & 0.012 & [0.007, 0.019] & 0.016 & [0.008, 0.035] & 0.013 & [0.008, 0.021] & 0.086 & [0.049, 0.122] & 0.121 & [0.076, 0.169] & 0.110 & [0.055, 0.587] \\
   &  & 256 & \textit{0.007} & [\textit{0.007}, \textit{0.007}] & \textit{0.007} & [\textit{0.007}, \textit{0.007}] & \textit{0.007} & [\textit{0.007}, \textit{0.007}] & 0.009 & [\textit{0.007}, 0.014] & 0.009 & [\textit{0.007}, 0.014] & 0.016 & [0.010, 0.025] \\
  \addlinespace[1.5pt]
   & SigLIP 2 & 1 & \textit{0.007} & [\textit{0.007}, \textit{0.007}] & \textit{0.007} & [\textit{0.007}, \textit{0.007}] & 0.008 & [\textit{0.007}, 0.015] & 0.010 & [\textit{0.007}, 0.016] & 0.010 & [\textit{0.007}, 0.016] & 0.236 & [0.133, 0.420] \\
   & ViT-B/16 & 16 & 0.011 & [0.007, 0.017] & 0.015 & [0.008, 0.027] & 0.013 & [0.008, 0.022] & 0.089 & [0.054, 0.145] & 0.109 & [0.066, 0.181] & 0.255 & [0.105, 0.511] \\
   &  & 256 & 0.008 & [\textit{0.007}, 0.013] & 0.008 & [\textit{0.007}, 0.013] & 0.012 & [0.007, 0.020] & 0.027 & [0.015, 0.045] & 0.027 & [0.015, 0.045] & 0.050 & [0.028, 0.084] \\
  \bottomrule
\end{tabular}
\endgroup

%% file: tables/radii_mem.tex
\begingroup
\footnotesize
\setlength{\tabcolsep}{2.5pt}
\begin{tabular}{@{}c@{\hspace{6pt}}l@{\hspace{8pt}}r@{\,}lr@{\,}lr@{\,}l@{\hspace{12pt}}r@{\,}lr@{\,}lr@{\,}l@{}}
  \toprule
    & & \multicolumn{6}{c}{$\epsilon = 10\%$} & \multicolumn{6}{c}{$\epsilon = 50\%$} \\
  \cmidrule(lr){3-8}\cmidrule(l){9-14}
  & \textbf{Encoder} & \multicolumn{2}{c}{$\eta^\ast$} & \multicolumn{2}{c}{$r_\eta$} & \multicolumn{2}{c}{$c_\eta$} & \multicolumn{2}{c}{$\eta^\ast$} & \multicolumn{2}{c}{$r_\eta$} & \multicolumn{2}{c}{$c_\eta$} \\
  \midrule
  \multirow{12}{*}{\rotatebox[origin=c]{90}{\textbf{\textit{Uniform}}}} & ResNet-50 & \multicolumn{12}{@{}l@{}}{{\color{black!25}\xleaders\hrule height 0.62ex depth -0.55ex\hskip 0pt plus 1filll}} \\
   & \quad pretrained & 0.152 & [0.080, 0.234] & 0.388 & [0.148, 0.814] & 0.152 & [0.083, 0.235] & 0.816 & [0.523, 1.222] & 6.495 & [1.184, 14.423] & 0.816 & [0.523, 1.222] \\
   & \quad true labels & 0.177 & [0.111, 0.265] & 0.453 & [0.232, 0.838] & 0.177 & [0.112, 0.265] & 0.888 & [0.597, 1.297] & 8.070 & [1.128, 24.378] & 0.888 & [0.602, 1.297] \\
   & \quad shuffled labels & 0.145 & [0.091, 0.221] & 0.219 & [0.118, 0.344] & 0.145 & [0.092, 0.221] & 0.583 & [0.383, 0.860] & 0.621 & [0.390, 0.990] & 0.614 & [0.428, 0.888] \\
   & ConvNeXt-T & \multicolumn{12}{@{}l@{}}{{\color{black!25}\xleaders\hrule height 0.62ex depth -0.55ex\hskip 0pt plus 1filll}} \\
   & \quad pretrained & 0.122 & [0.071, 0.206] & 0.122 & [0.071, 0.206] & 0.158 & [0.096, 0.257] & 0.403 & [0.223, 0.706] & 0.403 & [0.223, 0.718] & 0.619 & [0.364, 1.032] \\
   & \quad true labels & 0.157 & [0.108, 0.252] & 0.157 & [0.108, 0.252] & 0.221 & [0.144, 0.344] & 0.552 & [0.342, 0.946] & 0.552 & [0.342, 0.955] & 0.882 & [0.567, 1.362] \\
   & \quad shuffled labels & 0.230 & [0.124, 0.414] & 0.232 & [0.124, 0.437] & 0.273 & [0.160, 0.448] & 0.966 & [0.430, 1.644] & 1.318 & [0.432, 12.604] & 1.104 & [0.621, 1.720] \\
   & ViT-S/16 & \multicolumn{12}{@{}l@{}}{{\color{black!25}\xleaders\hrule height 0.62ex depth -0.55ex\hskip 0pt plus 1filll}} \\
   & \quad pretrained & 0.156 & [0.092, 0.257] & 0.181 & [0.096, 0.368] & 0.214 & [0.136, 0.330] & 0.699 & [0.432, 1.129] & 0.949 & [0.483, 2.200] & 0.842 & [0.554, 1.296] \\
   & \quad true labels & 0.225 & [0.135, 0.370] & 0.276 & [0.143, 0.504] & 0.287 & [0.205, 0.436] & 1.177 & [0.740, 1.721] & 2.098 & [0.956, 17.612] & 1.318 & [0.883, 1.853] \\
   & \quad shuffled labels & 0.776 & [0.493, 1.187] & 0.836 & [0.516, 1.316] & 0.938 & [0.649, 1.323] & 3.103 & [2.061, 4.704] & 3.658 & [2.284, 6.523] & 3.637 & [2.575, 5.113] \\
  \addlinespace[2pt]\arrayrulecolor{black!60}\cmidrule{1-14}\arrayrulecolor{black}\addlinespace[2pt]
  \multirow{12}{*}{\rotatebox[origin=c]{90}{\textbf{\textit{$k = 1$}}}} & ResNet-50 & \multicolumn{12}{@{}l@{}}{{\color{black!25}\xleaders\hrule height 0.62ex depth -0.55ex\hskip 0pt plus 1filll}} \\
   & \quad pretrained & \textit{0.007} & [\textit{0.007}, 0.008] & \textit{0.007} & [\textit{0.007}, 0.008] & 0.064 & [0.014, 0.146] & 0.034 & [0.012, 0.060] & 0.034 & [0.012, 0.060] & 2.490 & [0.589, 5.046] \\
   & \quad true labels & 0.008 & [\textit{0.007}, 0.013] & 0.008 & [\textit{0.007}, 0.013] & 0.179 & [0.092, 0.314] & 0.064 & [0.036, 0.099] & 0.064 & [0.036, 0.099] & 6.944 & [4.460, 10.511] \\
   & \quad shuffled labels & 0.007 & [\textit{0.007}, 0.012] & 0.007 & [\textit{0.007}, 0.012] & 0.228 & [0.111, 0.402] & 0.053 & [0.024, 0.089] & 0.053 & [0.024, 0.089] & 3.224 & [0.830, 5.389] \\
   & ConvNeXt-T & \multicolumn{12}{@{}l@{}}{{\color{black!25}\xleaders\hrule height 0.62ex depth -0.55ex\hskip 0pt plus 1filll}} \\
   & \quad pretrained & 0.027 & [0.014, 0.047] & 0.027 & [0.014, 0.047] & 0.107 & [0.052, 0.189] & 0.099 & [0.053, 0.172] & 0.099 & [0.053, 0.172] & 5.142 & [2.201, 10.037] \\
   & \quad true labels & 0.037 & [0.020, 0.056] & 0.037 & [0.021, 0.057] & 0.132 & [0.073, 0.208] & 0.127 & [0.072, 0.199] & 0.127 & [0.072, 0.199] & 4.840 & [2.627, 8.445] \\
   & \quad shuffled labels & 0.016 & [0.009, 0.028] & 0.016 & [0.009, 0.028] & 0.045 & [0.022, 0.109] & 0.055 & [0.032, 0.090] & 0.055 & [0.032, 0.090] & 4.005 & [0.900, 12.374] \\
   & ViT-S/16 & \multicolumn{12}{@{}l@{}}{{\color{black!25}\xleaders\hrule height 0.62ex depth -0.55ex\hskip 0pt plus 1filll}} \\
   & \quad pretrained & \textit{0.007} & [\textit{0.007}, \textit{0.007}] & \textit{0.007} & [\textit{0.007}, \textit{0.007}] & \textit{0.007} & [\textit{0.007}, 0.010] & \textit{0.007} & [\textit{0.007}, 0.013] & \textit{0.007} & [\textit{0.007}, 0.014] & 0.014 & [\textit{0.007}, 0.204] \\
   & \quad true labels & \textit{0.007} & [\textit{0.007}, \textit{0.007}] & \textit{0.007} & [\textit{0.007}, \textit{0.007}] & \textit{0.007} & [\textit{0.007}, \textit{0.007}] & \textit{0.007} & [\textit{0.007}, 0.008] & \textit{0.007} & [\textit{0.007}, 0.009] & 0.011 & [\textit{0.007}, 0.039] \\
   & \quad shuffled labels & 0.008 & [\textit{0.007}, 0.025] & 0.009 & [\textit{0.007}, 0.025] & 0.042 & [0.011, 0.116] & 0.030 & [0.010, 0.111] & 0.031 & [0.011, 0.111] & 1.380 & [0.098, 2.581] \\
  \addlinespace[2pt]\arrayrulecolor{black!60}\cmidrule{1-14}\arrayrulecolor{black}\addlinespace[2pt]
  \multirow{12}{*}{\rotatebox[origin=c]{90}{\textbf{\textit{$k = 16$}}}} & ResNet-50 & \multicolumn{12}{@{}l@{}}{{\color{black!25}\xleaders\hrule height 0.62ex depth -0.55ex\hskip 0pt plus 1filll}} \\
   & \quad pretrained & 0.007 & [\textit{0.007}, 0.012] & 0.007 & [\textit{0.007}, 0.012] & 0.059 & [0.029, 0.100] & 0.067 & [0.040, 0.100] & 0.068 & [0.040, 0.100] & 1.266 & [0.646, 2.099] \\
   & \quad true labels & 0.010 & [0.007, 0.015] & 0.011 & [0.007, 0.015] & 0.068 & [0.044, 0.103] & 0.099 & [0.068, 0.146] & 0.099 & [0.068, 0.146] & 1.067 & [0.711, 1.694] \\
   & \quad shuffled labels & 0.013 & [0.009, 0.018] & 0.013 & [0.009, 0.018] & 0.141 & [0.091, 0.206] & 0.116 & [0.087, 0.151] & 0.116 & [0.087, 0.151] & 1.590 & [1.137, 2.164] \\
   & ConvNeXt-T & \multicolumn{12}{@{}l@{}}{{\color{black!25}\xleaders\hrule height 0.62ex depth -0.55ex\hskip 0pt plus 1filll}} \\
   & \quad pretrained & 0.075 & [0.052, 0.107] & 0.078 & [0.056, 0.123] & 0.134 & [0.090, 0.210] & 0.327 & [0.231, 0.488] & 0.329 & [0.231, 0.494] & 2.849 & [1.710, 4.368] \\
   & \quad true labels & 0.095 & [0.070, 0.134] & 0.099 & [0.071, 0.141] & 0.186 & [0.130, 0.262] & 0.387 & [0.272, 0.536] & 0.387 & [0.272, 0.539] & 3.555 & [2.308, 5.118] \\
   & \quad shuffled labels & 0.049 & [0.037, 0.071] & 0.059 & [0.041, 0.086] & 0.068 & [0.045, 0.097] & 0.215 & [0.152, 0.296] & 0.221 & [0.159, 0.308] & 0.627 & [0.371, 1.104] \\
   & ViT-S/16 & \multicolumn{12}{@{}l@{}}{{\color{black!25}\xleaders\hrule height 0.62ex depth -0.55ex\hskip 0pt plus 1filll}} \\
   & \quad pretrained & 0.017 & [0.007, 0.049] & 0.019 & [0.007, 0.074] & 0.024 & [0.009, 0.068] & 0.372 & [0.189, 0.565] & 0.378 & [0.203, 0.572] & 1.271 & [0.733, 1.917] \\
   & \quad true labels & 0.024 & [0.008, 0.082] & 0.027 & [0.009, 0.113] & 0.035 & [0.011, 0.117] & 0.477 & [0.283, 0.736] & 0.485 & [0.290, 0.756] & 1.705 & [1.067, 2.511] \\
   & \quad shuffled labels & 0.130 & [0.082, 0.187] & 0.135 & [0.087, 0.205] & 0.248 & [0.149, 0.358] & 0.590 & [0.429, 0.759] & 0.590 & [0.429, 0.759] & 2.411 & [1.794, 3.176] \\
  \addlinespace[2pt]\arrayrulecolor{black!60}\cmidrule{1-14}\arrayrulecolor{black}\addlinespace[2pt]
  \multirow{12}{*}{\rotatebox[origin=c]{90}{\textbf{\textit{$k = 256$}}}} & ResNet-50 & \multicolumn{12}{@{}l@{}}{{\color{black!25}\xleaders\hrule height 0.62ex depth -0.55ex\hskip 0pt plus 1filll}} \\
   & \quad pretrained & 0.019 & [0.008, 0.030] & 0.046 & [0.015, 0.128] & 0.020 & [0.009, 0.031] & 0.195 & [0.095, 0.302] & 0.762 & [0.406, 1.369] & 0.195 & [0.096, 0.303] \\
   & \quad true labels & 0.029 & [0.018, 0.043] & 0.094 & [0.046, 0.220] & 0.029 & [0.018, 0.044] & 0.272 & [0.175, 0.387] & 1.545 & [0.916, 3.350] & 0.272 & [0.176, 0.387] \\
   & \quad shuffled labels & 0.022 & [0.014, 0.031] & 0.031 & [0.016, 0.069] & 0.022 & [0.014, 0.031] & 0.196 & [0.117, 0.279] & 0.554 & [0.144, 6.126] & 0.197 & [0.122, 0.279] \\
   & ConvNeXt-T & \multicolumn{12}{@{}l@{}}{{\color{black!25}\xleaders\hrule height 0.62ex depth -0.55ex\hskip 0pt plus 1filll}} \\
   & \quad pretrained & 0.075 & [0.046, 0.123] & 0.075 & [0.046, 0.124] & 0.095 & [0.062, 0.138] & 0.401 & [0.181, 0.757] & 0.519 & [0.181, 2.576] & 0.496 & [0.282, 0.789] \\
   & \quad true labels & 0.126 & [0.081, 0.187] & 0.128 & [0.081, 0.210] & 0.141 & [0.098, 0.205] & 0.730 & [0.431, 1.052] & 2.327 & [0.598, 3.655] & 0.759 & [0.504, 1.067] \\
   & \quad shuffled labels & 0.078 & [0.050, 0.124] & 0.079 & [0.050, 0.128] & 0.093 & [0.066, 0.135] & 0.401 & [0.212, 0.606] & 1.046 & [0.219, 2.019] & 0.431 & [0.284, 0.623] \\
   & ViT-S/16 & \multicolumn{12}{@{}l@{}}{{\color{black!25}\xleaders\hrule height 0.62ex depth -0.55ex\hskip 0pt plus 1filll}} \\
   & \quad pretrained & 0.013 & [0.007, 0.024] & 0.013 & [0.007, 0.024] & 0.019 & [0.010, 0.037] & 0.039 & [0.021, 0.073] & 0.039 & [0.021, 0.073] & 0.071 & [0.036, 0.142] \\
   & \quad true labels & 0.022 & [0.013, 0.038] & 0.022 & [0.013, 0.038] & 0.032 & [0.020, 0.050] & 0.067 & [0.040, 0.109] & 0.067 & [0.040, 0.109] & 0.125 & [0.074, 0.217] \\
   & \quad shuffled labels & 0.070 & [0.040, 0.104] & 0.070 & [0.040, 0.104] & 0.098 & [0.056, 0.144] & 0.249 & [0.133, 0.421] & 0.249 & [0.133, 0.421] & 0.499 & [0.276, 0.743] \\
  \bottomrule
\end{tabular}
\endgroup